\documentclass[runningheads]{llncs}
\usepackage{graphicx}
\usepackage{amsmath,amssymb} 
\usepackage{color}

\usepackage{lipsum}
\usepackage{graphics} 
\usepackage{epsfig} 
\usepackage{mathptmx} 
\usepackage{times} 
\usepackage{amsmath} 
\usepackage{amssymb}  
\usepackage{caption}
\usepackage{array,multirow}
\usepackage{verbatim}
\usepackage{gensymb} 
\usepackage{hyperref}
\usepackage{tabularx}
\usepackage{ltablex}
\usepackage{calrsfs}
\DeclareMathAlphabet{\pazocal}{OMS}{zplm}{m}{n}
\DeclareMathAlphabet\pazobfcal{OMS}{zplm}{b}{n}
\usepackage{booktabs} 

\keepXColumns
\begin{document}

\newcolumntype{Y}{>{\centering\arraybackslash}p}

\title{ENG: End-to-end Neural Geometry for Robust Depth and Pose Estimation using CNNs} 
\titlerunning{ENG: End-to-end Neural Geometry} 


\author{Thanuja Dharmasiri*, Andrew Spek*\thanks{*These authors contributed equally. This research was supported by the Australian Research Council Centre of Excellence for Robotic Vision (project number CE14010006).} \\
Tom Drummond}
\renewcommand\footnotemark{}
%

\authorrunning{T.Dharmairi, A.Spek, T.Drummond} 

\institute{Monash University, Australia\\
\email{\{thanuja.dharmasiri,andrew.spek,tom.drummond\}@monash.edu}}

\maketitle

\begin{abstract}

Recovering structure and motion parameters given a image pair or a sequence of images is a well studied problem in computer vision. This is often achieved by employing  Structure from Motion (SfM) or  Simultaneous Localization and Mapping (SLAM) algorithms based on the real-time requirements. Recently, with the advent of Convolutional Neural Networks (CNNs)  researchers have explored the possibility of using machine learning techniques to reconstruct the 3D structure of a scene and jointly predict the camera pose. In this work, we present a framework that achieves state-of-the-art performance on single image depth prediction for both indoor and outdoor scenes. The depth prediction system is then extended to predict optical flow and ultimately the camera pose and trained end-to-end. Our framework outperforms previous deep-learning based motion prediction approaches, and we also demonstrate that the state-of-the-art metric depths can be further improved using the knowledge of pose. 

\keywords{Depth, Optical Flow, Pose Prediction  \and Indoor and Outdoor Datasets}
\end{abstract}

\section{Introduction}

The importance of navigation and mapping to the fields of robotics and computer vision has only increased since its inception. Vision based navigation in particular is an extremely interesting field of research due to its discernible resemblance to human navigation and the wealth of information an image contains. Although creating a machine that understands structure and motion purely from RGB images is challenging,  the computer vision community has developed a plethora of algorithms to replicate useful aspects of human vision using a computer. Tracking and mapping remains an unsolved problem, with many popular approaches. Photometric based techniques rely on establishing correspondences across different viewpoints of the same scene and the matching points are then used to perform triangulation. Based on the density of the map, the field can be divided into dense \cite{Newcombe2011}, semi-dense \cite{engel2014lsd} and sparse\cite{mur2015orb} approaches, each comes with advantages and disadvantages.

Applying machine learning techniques to solve vision problems has been another popular area of research. Great advances have been made in the fields of image classification \cite{Krizhevsky,He,huang2017densely} and semantic segmentation \cite{Long,zhao2017pyramid} and this has led geometry based machine learning methods to follow suit. The massive growth in neural network driven research has largely been facilitated by the increased availability of low-cost high performance GPUs as well as the relative accessibility of machine learning frameworks such as Tensorflow and Caffe. 

In this work we draw from both machine learning approaches as well as SfM techniques to create a unified framework which is capable of predicting the depth of a scene and the motion parameters governing the camera motion between an image pair. We construct our framework incrementally where the network is first trained to predict depths given a single color image. Then a color image pair as well as their associated depth predictions are provided to a flow estimation network which produces an optical flow map along with an estimated measure of confidence in x and y motion. Finally, the pose estimation block utilises the outputs of the previous networks to estimate a motion vector corresponding to the logarithm of the Special Euclidean Transformation $\mathbb{SE}(3)$ in $\mathbb{R}^3$, which describes the relative camera motion from the first image to the second.

We summarise the contributions made in this paper as follows:
\begin{itemize}
\item We achieve state-of-the-art results for single image depth prediction on both NYUv2 (indoor) and KITTI (outdoor) datasets. [Section \ref{sec:results}: Table \ref{tab:nyu_depth} and Table \ref{tab:kitti_depth}]
\item We outperform previous camera motion prediction frameworks on both TUM and KITTI datasets. [Section \ref{sec:results}:  Table \ref{tab:rgbd_pose} and Table \ref{tab:kitti_pose}]
\item We  also  present  the  first  approach  to  use  a
neural network to predict the full information matrix which represents the confidence of our optical flow estimate.  
\end{itemize}

\section{Related Work}
Estimating motion and structure from two or more views is a well studied vision problem. In order to reconstruct the world and estimate camera motion, sparse feature based systems \cite{klein2007parallel,mur2015orb} compute correspondences through feature matching while the denser approaches\cite{engel2014lsd,Newcombe2011} rely on brightness constancy across multiple viewpoints.  In this work, we leverage CNNs to solve the aforementioned tasks and we summarize the existing works in the literature that are related to the ideas presented in this paper. 

\subsection{Single Image Depth Prediction} 

Predicting depth from a single RGB image using learning based approaches  has been explored even prior to the resurgence of CNNs. In \cite{saxena2006learning}, Saxena \emph{et al}. employed a Markov Random Field (MRF) to combine global and local image features.  Similar to our approach Eigen \emph{et al}. \cite{EigenNIPS} introduced a common CNN architecture capable of predicting depth maps for both indoor and outdoor environments. This concept was later extended to a multi-stage coarse to fine network by Eigen \emph{et al}. in \cite{Eigen}. Advances were made in the form of combining graphical models with CNNs \cite{Liu} to further improve the accuracy of depth maps, through the use of related geometric tasks \cite{dharmasiri2017joint} and by making architectural improvements specifically designed for depth prediction \cite{laina2016}. Kendall \emph{et al}. demonstrated that predicting depths and uncertainties improve the overall accuracy in \cite{kendall2017uncertainties}. While most of these methods demonstrated impressive results, explicit notion of geometry was not used during any stage of the pipeline which opened the way for geometry based depth prediction approaches.

In one of the earliest works to predict depth using geometry in an unsupervised fashion, Garg \emph{et al}. used the photometric difference between a stereo image pair, where the target image was synthesized using the predicted disparity and the known baseline\cite{garg2016unsupervised}. Left-right consistency was explicitly enforced in the unsupervised framework of Goddard \emph{et al}. \cite{godard2017unsupervised} as well as in the semi-supervised framework of Kuznietsov \emph{et al}.\cite{kuznietsov2017semi}, which is a technique we also found to be beneficial during training on sparse ground truth data.

\subsection{Optical Flow Prediction}

An early work in optical flow prediction using CNNs was \cite{fischer2015flownet}. This was later extended by Ilg \emph{et al}. to FlowNet 2.0 \cite{ilg2017flownet} which included stacked FlowNets \cite{fischer2015flownet} as well as warping layers. Ranjan and Black proposed a spatial pyramid based optical flow prediction network \cite{ranjan2017optical}.  More recently, Sun \emph{et al}. proposed a framework which uses the principles from geometry based flow estimation techniques such as image pyramid, warping and cost volumes in \cite{sun2017pwc}. As our end goal revolves around predicting camera pose, it becomes necessary to isolate the flow that was caused purely from camera motion, in order to achieve this we extend upon these previous works to predict both the optical flow and the associated information matrix of the flow. Although not in a CNN context \cite{wannenwetsch2017probflow} showed the usefulness of estimating flow and uncertainty.

\subsection{Pose Estimation}

CNNs have been successfully used to estimate various components of a Structure from Motion pipeline. Earlier works focused on learning discriminative image based features suitable for ego-motion estimation \cite{agrawal2015learning,jayaraman2015learning}. Yi \emph{et al}.\cite{Yi16b} showed a full feature detection framework can be implemented using deep neural networks. Rad and Lepetit in BB8\cite{Rad17c} showed the pose of objects can be predicted even under partial occlusion and highlighted the increased difficulty of predicting 3D quantities over 2D quantities. Kendall and Cipolla demonstrated that camera pose prediction from a single image catered for relocalization scenarios \cite{kendall2016modelling}. 

However, each of the above works lack a representation of structure as they do not explicitly predict depths. Our work is more closely related to that of Zhou \emph{et al}. \cite{zhou2017unsupervised} and Ummenhofer \emph{et al}. \cite{UZUMIDB17} and their frameworks SfM-Learner and DeMoN. Both of these approaches also predict a single confidence map in contrast to ours which estimates the confidence in x and y directions separately. Since our framework predicts metric depths in comparison to theirs we are able produce far more accurate visual odometry and combat against scale drift. CNN SLAM by Tateno \emph{et al.} \cite{tateno2017cnn} incorporated depth predictions of \cite{laina2016} into a SLAM framework. Our method performs competitively with CNN-SLAM as well as ORB-SLAM\cite{murORB2} and LSD-SLAM\cite{engel2014lsd} which have the added advantage of performing loop closures and local/global bundle adjustments despite solely computing sequential frame-to-frame alignments. 

\section{Method}

\subsection{Network Architecture}
\begin{figure}[h!]
	\begin{center}
    	\includegraphics[width=\linewidth]{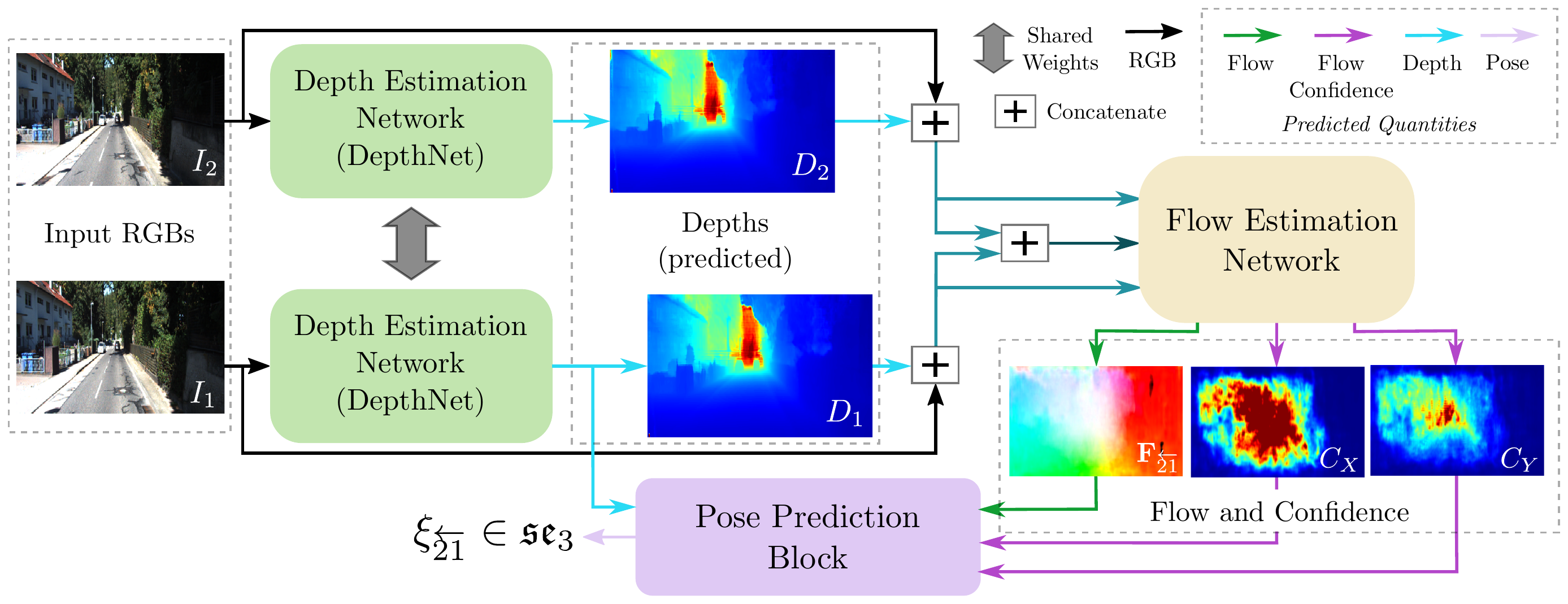}
		\caption{Overview of our system full pipeline. Please note that we use the notation $\protect \overleftarrow{ij}$ to indicate from $j$ to $i$}
		\label{fig:network_arch}
	\end{center}
    \vskip -3em
\end{figure}

The overall architecture consists of 3 main subsystems in the form of a depth, flow and camera pose network. A large percentage of the model capacity is invested in to the depth prediction component for two reasons. Firstly, the output of the depth network also serves as an additional input to the other subsystems. Secondly, we wanted to achieve superior depths for indoor and outdoor environments using a common architecture \footnote{Although there are separate models for indoor and outdoor scenes the underlying architecture is common.}. In order to preserve space and to provide an overall understanding of the data flow a high level diagram of the network is shown in Figure \ref{fig:network_arch}. An expanded architecture with layer definitions for each of the subsystems is included in the supplementary materials. 
\subsection{Depth Prediction}
The depth prediction network consists of an encoder and a decoder module. The encoder network is largely based on the DenseNet161 architecture described in \cite{huang2017densely}. In particular we use the variant pre-trained on ImageNet \cite{ILSVRC15} and slightly increase the receptive field of the pooling layers. As the original input is down-sampled 4 times by the encoder, during the decoding stage the feature maps are up-sampled back 4 times to make the model fully convolutional. We employ skip connections in order to re-introduce the finer details lost during pooling. Since the first down-sampling operation is done at a very early stage of the pipeline and closely resemble the image features, these activations are not reused inside the decoder. Up-project blocks are used to perform up-sampling in our network, which provide better depth maps compared to de-convolutional layers as shown in \cite{laina2016}.

Due to the availability of dense ground truth data for indoor datasets (e.g NYUv2 \cite{Silberman}, RGB-D\cite{sturm12iros}) this network can be directly utilised to perform supervised learning. Unfortunately, the ground truth data for the outdoor datasets (KITTI) are much sparser and meant we had to incorporate a semi-supervised learning approach in order to provide a strong training signal. Therefore, during training on KITTI, we use a Siamese version of the depth network with complete weight-sharing, and enforce photometric consistency between the left-right image pairs through an additional loss function. This is similar to the previous approaches \cite{garg2016unsupervised,kuznietsov2017semi} and is only required during the training stage, during inference only a single input image is required to perform depth estimation using our network. 

\subsection{Flow Prediction}

The flow network provides an estimation of the optical flow along with the associated confidences given an image pair. These outputs combined with predicted depths allow us to predict the camera pose. As part of our ablation studies we integrated the flow predictions of \cite{ilg2017flownet} with our depths, however, the main limitation of this approach was the lack of a mechanism to filter out the dynamic objects which are abundant in outdoor environments.  This was solved by estimating confidence, specifically the information matrix in addition to the optical flow. More concretly, for each pixel our flow network predicts 5 quantities, the optical flow $\mathbf{F} = [\Delta u, \Delta v]^T$ in the x and y direction, and the quantities $\hat{\alpha}$, $\hat{\gamma}$ and $\hat{\beta}$, which are required to compute the information matrix $(\pazobfcal{I})$ or the inverse of the covariance matrix as shown below.
\begin{equation}
\mathbf{\pazobfcal{I}}=
\left[\begin{array}{cccc}
C_x & C_{xy} \\
C_{xy} & C_y \\
\end{array}\right], \quad\quad C_x =e^{\hat{\alpha}}, C_y = e^{\hat{\gamma}}, C_{xy} = e^{\frac{\hat{\gamma} + \hat{\alpha}}{2}} \tanh(\hat{\beta}).
\label{eqn:infomat}
\end{equation}

This parametrisation guarantees $\pazobfcal{I}$ is positive-definite and can be used to parametrise any $2 \times 2$ information matrix. We found that the gradients are much more stable compared to predicting the information matrix directly as the determinant of the matrix is always greater than zero since $\tanh(\hat\beta) = \pm1$ only when $ \hat{\beta} \to \pm\infty$.

With respect to the architecuture we borrow elements from FlowNet \cite{fischer2015flownet} as well as FlowNet 2.0 \cite{ilg2017flownet}. As mentioned in \cite{ilg2017flownet}, FlowNet 2.0 was unable to reliably estimate small motions, which we address with two key changes. Firstly, our flow network takes the predicted depth map as an input, allowing the network to learn the relationship between depth and flow explicitly, including that closer objects appear to move more compared to the objects that are further away from the camera. Secondly, we use ``warp-concatenation'', where coarse flow estimates are used to warp the CNN features during the decoder stage. This appears to resolve small motions more effectively particularly on the TUM \cite{sturm12iros} dataset.

\begin{figure}[h!]
	\begin{center}
    	\includegraphics[width=.9\linewidth]{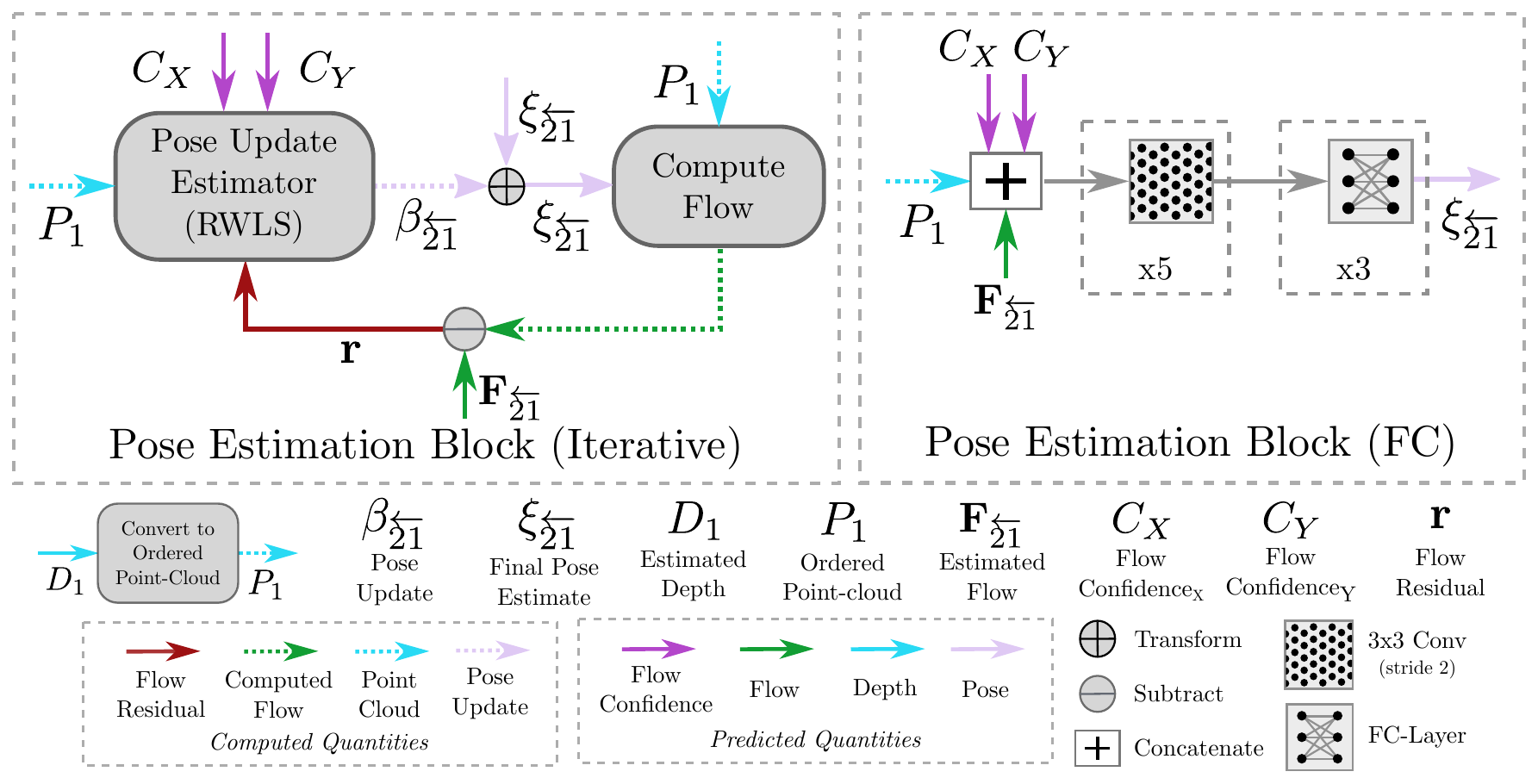}
		\caption{We detail the two approaches we took to estimating the relative pose alignment between adjacent frames (best viewed in colour). \emph{Left} shows the iterative approach we took, that incorporates a re-weighted least-squares solver (RWLS) into a pose estimation loop. \emph{Right} shows our fully-connected (FC) approach, which incorporates a succession of $3\times3$ strided convolutions, followed by several FC layers.}
		\label{fig:pose_net}
	\end{center}
\end{figure}

\subsection{Pose Estimation}

We take two approaches to pose estimation, shown in Figure \ref{fig:pose_net}, an iterative and a fully-connected(FC). This contrasts the ability of a neural network to estimate using the available information, and the simplicity of a standard computer vision approach using the available predicted quantities. We use FC layers to provide the network with as wide a receptive field as possible, to compare more equivalently against using the inferred quantities in the iterative approach.  

\subsubsection*{Iterative}
\label{subsec:iterative}
This approach uses a more conventional method for computing relative pose estimates. We use a standard re-weighted least squares solver based on the residual flow, given an estimate of the relative transformation. More concretely we attempt to minimise the following error function with respect to the relative transformation parameters ($\xi_{\overleftarrow{21}} \in \mathbb{R}^6$)
\begin{equation}
\mathbf{e} = \sum_{i=1}^{N} ||(\mathbf{x}_i - \text{T}_{\overleftarrow{21}}\mathbf{x}_i)_{[u,v]} - \mathbf{F}_{\overleftarrow{21}}(\mathbf{u}_i)||_2 = \sum_{i=1}^{N} ||\mathbf{F}^+_{\overleftarrow{21}}(\mathbf{u}_i) - \mathbf{F}_{\overleftarrow{21}}(\mathbf{u}_i)||_2 = \sum_{i=1}^{N} \mathbf{r}_i^2,
\label{eq:error_function_iterative}
\end{equation}
where $\mathbf{e} \in \mathbb{R}^2$ is the total residual flow in normalised camera coordinates, the subscript $\phantom{}_{[u,v]}$ indicates only the first two dimensions of the vector are used, $\mathbf{x} \in P_1$ is the $i^{th}$ inverse depth coordinate $\mathbf{x} = \left[\begin{array}{cccc}
u & v & 1 & q \\
\end{array}\right]^T$ of an ordered point cloud ($P_1$), $\mathbf{F}_{\overleftarrow{21}}(\mathbf{u}_i)$ and $\mathbf{F}^+_{\overleftarrow{21}}(\mathbf{u}_i)$ are the $i^{th}$ predicted flow and estimated flow respectively, and $\mathbf{u}_i=\left[\begin{array}{cc} x & y\\ \end{array} \right]_i^T$ is the $i^{th}$ pixel coordinate. $\text{T}_{\overleftarrow{21}}\in \mathbb{SE}(3)$ is the current transformation estimate, and can be expressed by the matrix exponential as $\text{T}_{\overleftarrow{21}} = e^{\sum_{j=0}^{6}\alpha_j \mathbf{G}_j}$, where $\alpha_j \in \xi_{\overleftarrow{21}}$ is the $j^{th}$ component of the motion vector $\xi_{\overleftarrow{21}} \in \mathbb{R}^6$, which is a member of the Lie-algebra $\mathfrak{se}_3$, and $\mathbf{G}_j$ is the generator matrix corresponding to the relevant motion parameter. As This pipeline is implemented in Tensorflow \cite{abadi2016tensorflow} it allows us to train the network end to end. Please see the supplementary material for a more detailed explanation.

\subsubsection*{Fully-Connected}

Similar to Zhou \emph{et al}.\cite{zhou2017unsupervised} and Ummenhofer \emph{et al}. \cite{UZUMIDB17} we also constructed a fully connected layer based pose estimation network. This network utilises 3 stacked fully connected layers and uses the same inputs as our iterative method. While we outperform the pose estimation benchmarks of \cite{zhou2017unsupervised} and \cite{UZUMIDB17} using this network the iterative network is our recommended approach due to its close resemblance to conventional geometry based techniques. 
\label{subsec:fullyconnected}
\subsection{Loss Functions}

\subsection*{Depth Losses}
For supervised training on indoor and outdoor datasets we use a reverse Huber loss function \cite{laina2016} defined by
\begin{equation}
  \pazocal{L}_\text{B}(D_i,D^*_i)=\begin{cases}
|D_i -D^*_i| & |D_i -D^*_i|<c, \\
((D_i -D^*_i)^2+c^2)/2c & |D_i -D^*_i| > c,
  \end{cases}
\end{equation}
where $c = \frac{1}{5} max(D_i -D^*_i)$, and $D_i = D(\mathbf{u}_i)$ and $D^*_i = D^*(\mathbf{u}_i)$ represent the $i^{th}$ predicted and the ground truth depth respectively. For the KITTI dataset we employed an additional photometric loss during training as the ground truth is highly sparse. This unsupervised loss term enforces left-right consistency between stereo pairs, defined by
\begin{equation}
\begin{split}
\pazocal{L}_\text{C} = \frac{1}{n}\sum_{i=1}^{n} |I_L(\mathbf{u}_i) - I_R(\pi(\text{K} \text{T}_{\overleftarrow{RL}} \pi^{-1}(D^L_i, \mathbf{u}_i))| \\ 
+ \frac{1}{n}\sum_{i=1}^{n} |I_R(\mathbf{u}_i) - I_L(\pi(\text{K} \text{T}_{\overleftarrow{LR}} \pi^{-1}(D^R_i, \mathbf{u}_i))|, 
\end{split}
\end{equation}
where $I_L$ and $I_R$ are the left and right images and $D^L_i$ and $D^R_i$ are their corresponding depth maps, $\pi(x)=((x_0/x_2),(x_1/x_2))^T$ is a normalisation function where $x \in \mathbb{R}^3$, $\text{K}$ is the camera intrinsic matrix, $\pi^{-1}(D,\mathbf{u}) = D\text{K}^{-1}(\mathbf{u})$ is the transformation from pixel to  camera coordinates, and $\text{T}_{\overleftarrow{RL}} \in \mathbb{SE}(3)$ and $\text{T}_{\overleftarrow{LR}} \in \mathbb{SE}(3)$ define the relative transformation matrices from left-to-right and right-to-left respectively. In this case the rotation is assumed to be the identity and the matrices purely translate in the x-direction. Additionally, we use a smoothness term defined by
\begin{equation}
\pazocal{L}_\text{S} = \frac{1}{n}\sum_{i=1}^{n}\left(|\nabla_x D_i| + |\nabla_y D_i|\right),
\end{equation} 
where $\nabla_x$ and $\nabla_y$ are the horizontal and vertical gradients of the predicted depth. This provides qualitatively better depths as well as faster convergence. The final loss function used to train KITTI depths is given by
\begin{equation}
\pazocal{L}_{ss} =  \lambda_1 \pazocal{L}_\text{B} + \lambda_2  \pazocal{L}_\text{C} + \lambda_3  \pazocal{L}_\text{{S}}
\label{eqn:full_kitti_loss}, \quad\quad  \text{$\lambda_1 = 2$, $\lambda_2 = 1$ , $\lambda_3 = e^{-4}$ },
\end{equation}
where $\pazocal{L}_\text{B}$ and $\pazocal{L}_\text{S}$ are computed on both left and right images separately.

\subsection*{Flow Loss}
The probability distribution of multivariate Gaussian in 2D can be defined as follows. 
\begin{equation}
	p(\mathbf{x}|\mathbf{\mu},\pazobfcal{I}) =  ((|\pazobfcal{I}|^\frac{1}{2})/(2\pi))e^{-\frac{1}{2}(\mathbf{x}-\mathbf{\mu})^T\pazobfcal{I}(\mathbf{x}-\mathbf{\mu})},
 \label{ref:smoothness}
\end{equation}
where $\pazobfcal{I} = \Sigma^{-1}$ is the information matrix or inverse covariance matrix $\Sigma^{-1}$. The flow loss $\pazocal{L}_{F}$ criterion can now be defined by 
\begin{equation}
	\pazocal{L}_{F} =  \frac{1}{2}({(\mathbf{F}_{\overleftarrow{21}}-\mathbf{F}^*_{\overleftarrow{21}})^T\pazobfcal{I}(\mathbf{F}_{\overleftarrow{21}}-\mathbf{F}^*_{\overleftarrow{21}})} - \text{log}(\pazobfcal{|I|})),
 \label{ref:flow_loss}
\end{equation}
where $\mathbf{F}_{\overleftarrow{21}}$ is the predicted flow, and $\mathbf{F}^*_{\overleftarrow{21}}$ is the ground truth flow. This optimises by maximising the log-likelihood of the probability distribution over the residual flow error. 

\subsection*{Pose Loss}

Given two input images $I_{1}$, $I_{2}$, the predicted depth map $D_1$ of  $I_{1}$ and the predicted relative pose $\xi_{\overleftarrow{21}} \in \mathbb{R}^6$ the unsupervised loss $\pazocal{L}_\text{U}$ and pose loss $\pazocal{L}_\text{P}$ can be defined as
\begin{align}
\pazocal{L}_\text{U}& = \frac{1}{n}\sum_{i=1}^{n} |I_1(\mathbf{u}_i) - I_2(\pi(\text{K} \text{T}_{\overleftarrow{21}} \pi^{-1}(D_1, \mathbf{u}_i))|, \text{and}\\ 
\pazocal{L}_\text{P} &= ||\xi_{\overleftarrow{21}}-\log_e(\text{T}^*_{\overleftarrow{21}})||_2 = ||\xi_{\overleftarrow{21}}-\xi^*_{\overleftarrow{21}})||_2,
\end{align}
where $\log_e(\text{T})$ maps a transformation $\text{T}$ from the Lie-group $\mathbb{SE}(3)$ to the Lie-algebra $\mathfrak{se}_3$, such that $\log_e(\text{T}) \in \mathbb{R}^6$ can be represented by its constituent motion parameters, and $\xi^*_{\overleftarrow{21}}$ is the ground truth relative pose parameters.


\subsection{Training Regime}

We train our network end-to-end on NYUv2 \cite{Silberman}, TUM\cite{sturm12iros} and KITTI\cite{Geiger2013IJRR} datasets. We use the standard test/train split for NYUv2 and KITTI and define our scene split for TUM. It is worth mentioning that the amount of training data we used is radically reduced compared to \cite{zhou2017unsupervised} and \cite{UZUMIDB17}. More concretly, for NYUv2 we use $\approx 3\%$ of the full dataset, for KITTI $\approx25\%$. We use the Adam optimiser \cite{kingma2014adam} with an initial learning rate of 1e-4 for all experiments and chose Tensorflow \cite{abadi2016tensorflow} as the learning framework and train using an NVIDIA-DGX1. We provide a detailed training schedule and breakdown in the supplementary material.

\section{Results}
\label{sec:results}
In this section we summarise the single-image depth prediction and relative pose estimation performance of our system on several popular machine learning and SLAM datasets.  We also investigate the effect of using alternative optical flow estimates from \cite{ilg2017flownet} and \cite{revaud2015epicflow} in our pose estimation pipeline as an ablation study. The entire model contains $\approx$ 130M parameters. Our depth estimator runs at 5fps on an NVIDIA GTX 1080Ti, while other sub-networks run at $\approx$30fps.

\subsection{Depth Estimation}
We summarise the results of evaluating our single-image depth estimation of the datasets NYUv2\cite{Silberman}, RGB-D\cite{sturm12iros} and KITTI\cite{Geiger2013IJRR} in Tables \ref{tab:nyu_depth}, \ref{tab:kitti_depth} and \ref{tab:rgbd_depth} respectively using the established metrics of \cite{EigenNIPS}.

We train \emph{Ours(baseline)} model to showcase the improvement we get by purely using the depth loss. This is then extended to use the full end-to-end training loss (depth + flow + pose losses) in the \emph{Ours(full)} model which demonstrates a consistent improvement across all datasets. Most notably in Tables \ref{tab:kitti_depth} and \ref{tab:rgbd_depth} for which ground truth pose data was available for training. This validates our approach for improving single image depth estimation performance, and demonstrates a network can be improved by enforcing more geometric priors on the loss functions. We would like to mention that the improvement we gain from \emph{Ours(baseline)} to \emph{Ours(full)} is purely due to the novel combined loss terms as the flow and pose sub networks do not increase the model capacity of the depth subnet itself.

\begin{table}[h!]
\centering
\caption{The performance of several approaches evaluated on single-image depth estimation using  the standard testset of NYUv2\cite{Silberman} proposed in \cite{Eigen}.}
\begin{tabular}{cY{1.5cm}Y{1.5cm}Y{1.5cm}Y{1.5cm}Y{1.5cm}Y{1.5cm}}
\toprule
  \multirow{2}{*}{Method} & \multicolumn{3}{c}{\textit{lower better}} & \multicolumn{3}{c}{\textit{higher better}} \\
  &$\text{RMS}_{lin}$ & $\text{RMS}_{ln}$ & $\text{Rel}_{abs}$ & $\delta$ & $\delta^2$ & $\delta^3$ \\  \midrule
  $\text{Eigen}_{vgg}$ \cite{Eigen}  & 0.641 & 0.214 & 0.16 & 76.9\% & 95.0\% & 98.8\% \\
  Laina \emph{et al}.\cite{laina2016} & 0.573 & 0.195 & 0.13 & 81.1\% & 95.3\% & 98.8\% \\
  Kendall \emph{et al}. \cite{kendall2017uncertainties} & 0.506 & - & \textbf{0.110} & 81.7\%	& 95.9\% &	98.9\% \\ \midrule
  Ours (baseline) & 0.487 & 0.164 & 0.113 & {86.7\%} & {97.7\%} & {99.4\%} \\
  Ours (full) & \textbf{0.478} & \textbf{0.161} & 0.111 & \textbf{87.2\% }& \textbf{97.8\%} & \textbf{99.5\%} \\
\end{tabular}
\label{tab:nyu_depth}
\end{table}

\begin{table}[h!]
\centering
\caption{The performance of previous state-of-the-art approaches evaluated on the standard testset of the KITTI dataset \cite{Geiger2013IJRR}.}
\label{tab:kitti_depth}
\begin{tabular}{Y{0.5cm}cY{1.3cm}Y{1.3cm}Y{1.3cm}Y{1.3cm}Y{1.3cm}Y{1.3cm}}
  \toprule
  \multirow{2}{*}{Cap}& \multirow{2}{*}{Method} & \multicolumn{3}{c}{\textit{lower better}} & \multicolumn{3}{c}{\textit{higher better}} \\
  &  &$\text{RMS}_{lin}$ & $\text{RMS}_{ln}$ & $\text{Rel}_{abs}$ & $\delta$ & $\delta^2$ & $\delta^3$ \\ \midrule 
  \multirow{5}{*}{\rotatebox{90}{0-80m}} & Zhou \emph{et al.}\cite{zhou2017unsupervised} & 6.856 & 0.283 & 0.208 & 67.8\% & 88.5\% & 95.7\% \\
  & Godard \emph{et al}.\cite{godard2017unsupervised} & 4.935 & 0.206 & 0.141 & 86.1\% & 94.9\% & 97.6\% \\
  & Kuznietsov \emph{et al}. \cite{kuznietsov2017semi} & 4.621 & 0.189 & 0.113 & 86.2\% & 96.0\% & 98.6\% \\
  & Ours (baseline) & 4.394 & 0.178 & 0.095 & 89.4\% & 96.6\% & 98.6\% \\
  & Ours (full) & \textbf{4.301} & \textbf{0.173} & \textbf{0.096} & \textbf{89.5\%} & \textbf{96.8\%} & \textbf{98.7\%} \\ \midrule
  \multirow{6}{*}{\rotatebox{90}{0-50m}} & Zhou \emph{et al.}\cite{zhou2017unsupervised} &5.181 & 0.264 & 0.201 & 69.6\% & 90.0\% & 96.6\% \\
  & Garg \emph{et al}. \cite{garg2016unsupervised} & 5.104 & 0.273 & 0.169 & 74.0\% & 90.4\% & 96.2\% \\
  & Godard \emph{et al}. \cite{godard2017unsupervised}& 3.729 & 0.194 & 0.108 & 87.3\% & 95.4\% & 97.9\% \\
  & Kuznietsov \emph{et al}.\cite{kuznietsov2017semi}& 3.518 & 0.179 & 0.108 & 87.5\% & 96.4\% & 98.8\% \\
  & Ours(baseline) & 3.359 & 0.168 & 0.092 & 90.5\% & 97.0\% & 98.8\% \\ 
  & Ours(full) & \textbf{3.284} & \textbf{0.164} & \textbf{0.092} & \textbf{90.6\%} & \textbf{97.1\%} & \textbf{98.9\%}\\
\end{tabular}
\end{table}
\begin{figure}[h!]
\centering
    \includegraphics[width=0.9\linewidth]{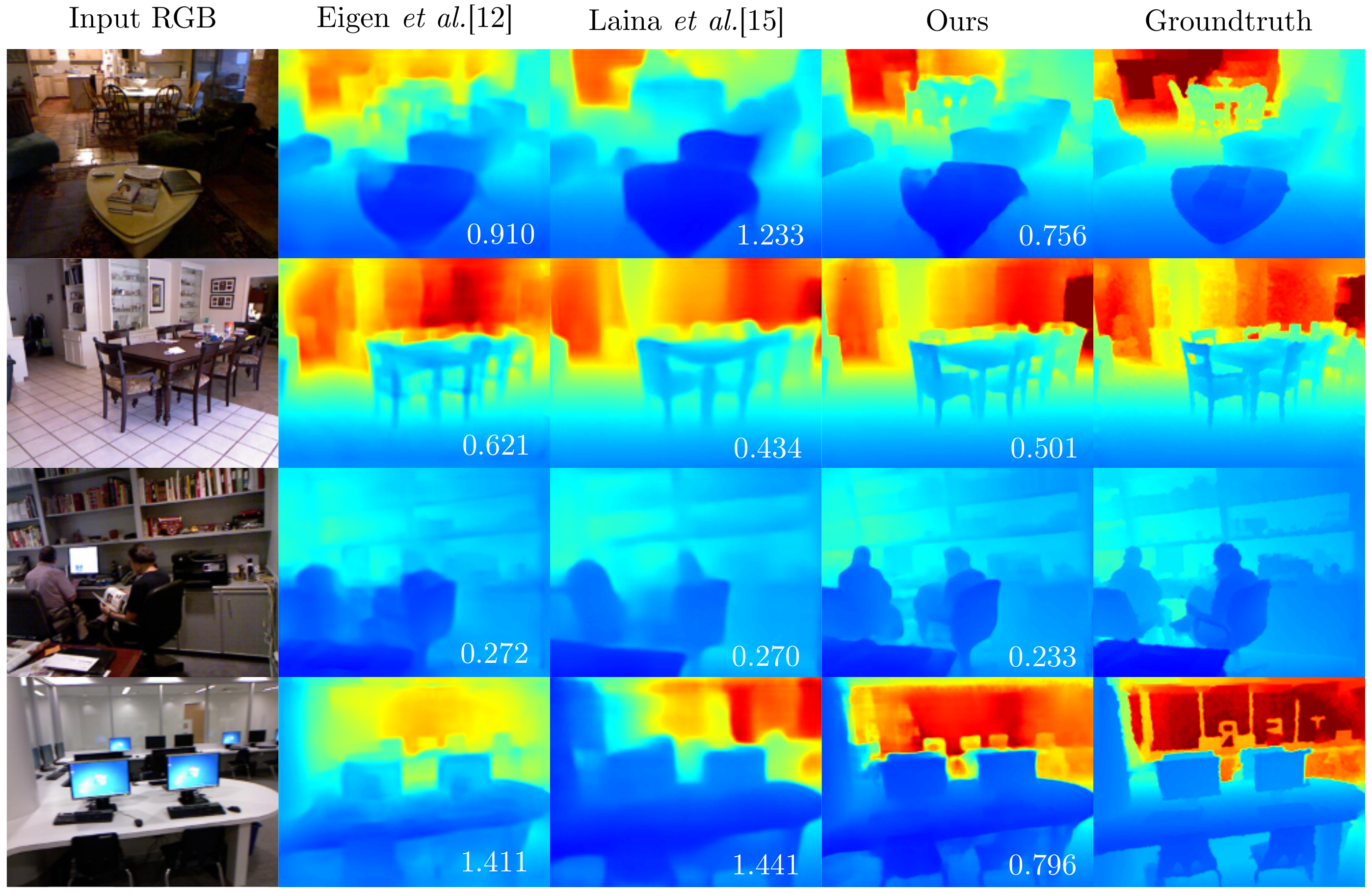}
	\caption{Resulting single image depth estimation for several approaches and ours against the ground truth on the dataset NYUv2\cite{Silberman}. The RMSE for each prediction is included}
	\label{fig:nyu_qualitative}
\end{figure}
\begin{table}[h!]
\centering
\caption{The performance of previous state-of-the-art approaches on a randomly selected subset of the frames from the RGB-D dataset \cite{sturm12iros}. We post separate entries for DeMoN(est) and DeMoN(gt), former is scaled by the estimated scale of their system while the latter is scaled by the median groundtruth depth.}
\begin{tabular}{cY{1.3cm}Y{1.3cm}Y{1.3cm}Y{1.3cm}Y{1.3cm}Y{1.3cm}Y{1.3cm}}
  \toprule
  \multirow{2}{*}{Method} & \multicolumn{4}{c}{\textit{lower better}} & \multicolumn{3}{c}{\textit{higher better}} \\
  &$\text{RMS}_{lin}$ & $\text{RMS}_{log}$ & $\text{Rel}_{abs}$ &  $\text{Rel}_{sqr}$ & $\delta$ & $\delta^2$ & $\delta^3$ \\ \midrule 
  Laina \emph{et al}.\cite{laina2016} & 1.275 & 0.481 & 0.189 & 0.371 & 75.3\% & 89.1\% & 91.8\% \\
  DeMoN(est)\cite{UZUMIDB17} & 2.980 & 0.910 & 1.413 & 5.109 & 21.0\%	& 36.6\% & 48.9\% \\
  DeMoN(gt)\cite{UZUMIDB17} & 1.584 & 0.555 & 0.301 & 0.581 & 52.7\% & 70.7\% & 80.7\% \\ \midrule
  Ours(baseline) & 1.068 & 0.353 & 0.128 & 0.236 & 86.9\% & 92.2\% & 93.5\% \\
  Ours(full) & \textbf{0.996} & \textbf{0.329} & \textbf{0.108} & \textbf{0.194} & \textbf{90.3\%}& \textbf{93.6\%} & \textbf{94.5\%} \\
\end{tabular}
\label{tab:rgbd_depth}
\end{table}
\begin{figure}[h!]
	\centering
    \includegraphics[width=0.9\linewidth]{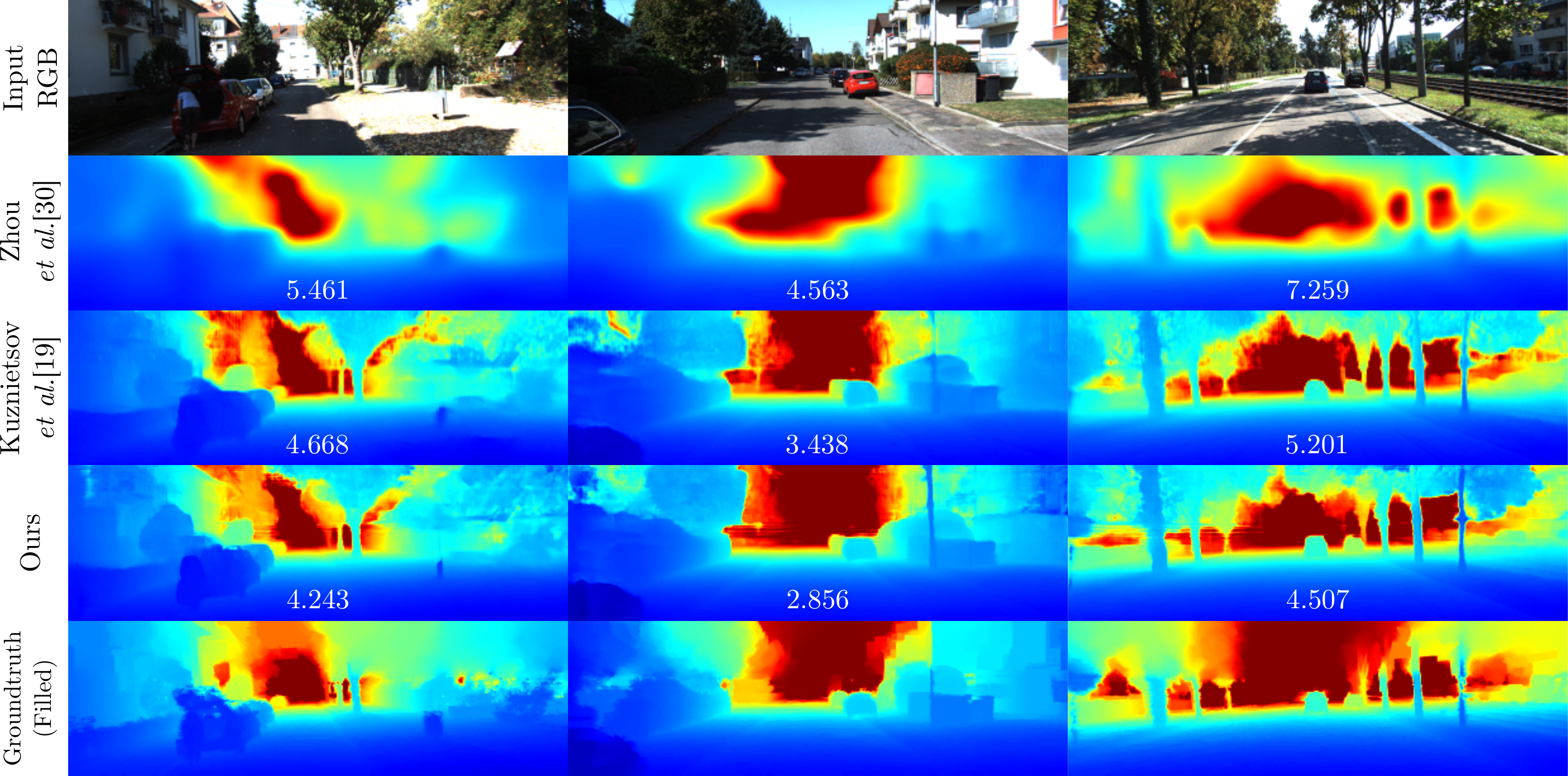}
	\caption{The resulting single image depth estimation for several approaches including Zhou \textit{et al.}\cite{zhou2017unsupervised}(SfM-Learner), Kuznietsov \textit{et al.}\cite{kuznietsov2017semi} and Ours against a ground truth filled using \cite{levin2004colorization} on the testset of the KITTI dataset \cite{Geiger2013IJRR}. We include the RMSE values for each methods prediction. Filled depths are included for visualisation purposes during evaluation the predictions are evaluated against the sparse velodyne ground truth data}
	\label{fig:kitti_qualitative}
\end{figure}

Additionally we include qualitative results for NYUv2\cite{Silberman} and KITTI\cite{Geiger2013IJRR} in Figure \ref{fig:nyu_qualitative} and \ref{fig:kitti_qualitative} respectively. Each of which illustrates a noticeable improvement over previous methods. We also demonstrate that the improvement is beyond the numbers, as our approach generates more convincing depths even when the RMSE may be higher, as is the case in the second row of Figure \ref{fig:nyu_qualitative}, where \cite{laina2016} computes a lower RMSE. More impressive still are the results in Figure \ref{fig:kitti_qualitative}, where we compare against previous approaches that are both trained on much larger training sets than our own and still show noticeable qualitative and quantitative improvements.

\subsection{Pose Estimation}

To demonstrate the ability of our approach to perform accurate relative pose estimation, we compare our approach on several unseen sequences from the datasets for which ground-truth poses were available. To quantitatively evaluate the trajectories we use the absolute trajectory error (ATE) and the relative pose error (RPE) as proposed in \cite{sturm12iros}. To mitigate the effect of scale-drift on these quantities we scale all poses to the groundtruth associated poses during evaluation. By using both metrics it provides an estimate of the consistency of each pose estimation approach. We summarise the results of this quantitative analysis for KITTI\cite{Geiger2013IJRR} in Table \ref{tab:kitti_pose} and for RGB-D\cite{sturm12iros} in Table \ref{tab:rgbd_pose}. We include comparisons of the performance against other state-of-the-art pose estimation networks namely SFM-Learner\cite{zhou2017unsupervised} and DeMoN\cite{UZUMIDB17}. Additionally we include results from current state-of-the-art SLAM systems also, namely ORB-SLAM2\cite{mur2015orb} and LSD-SLAM\cite{engel2014lsd}.

\begin{figure}[h!]
	\begin{center}
    	\includegraphics[width=0.90\linewidth]{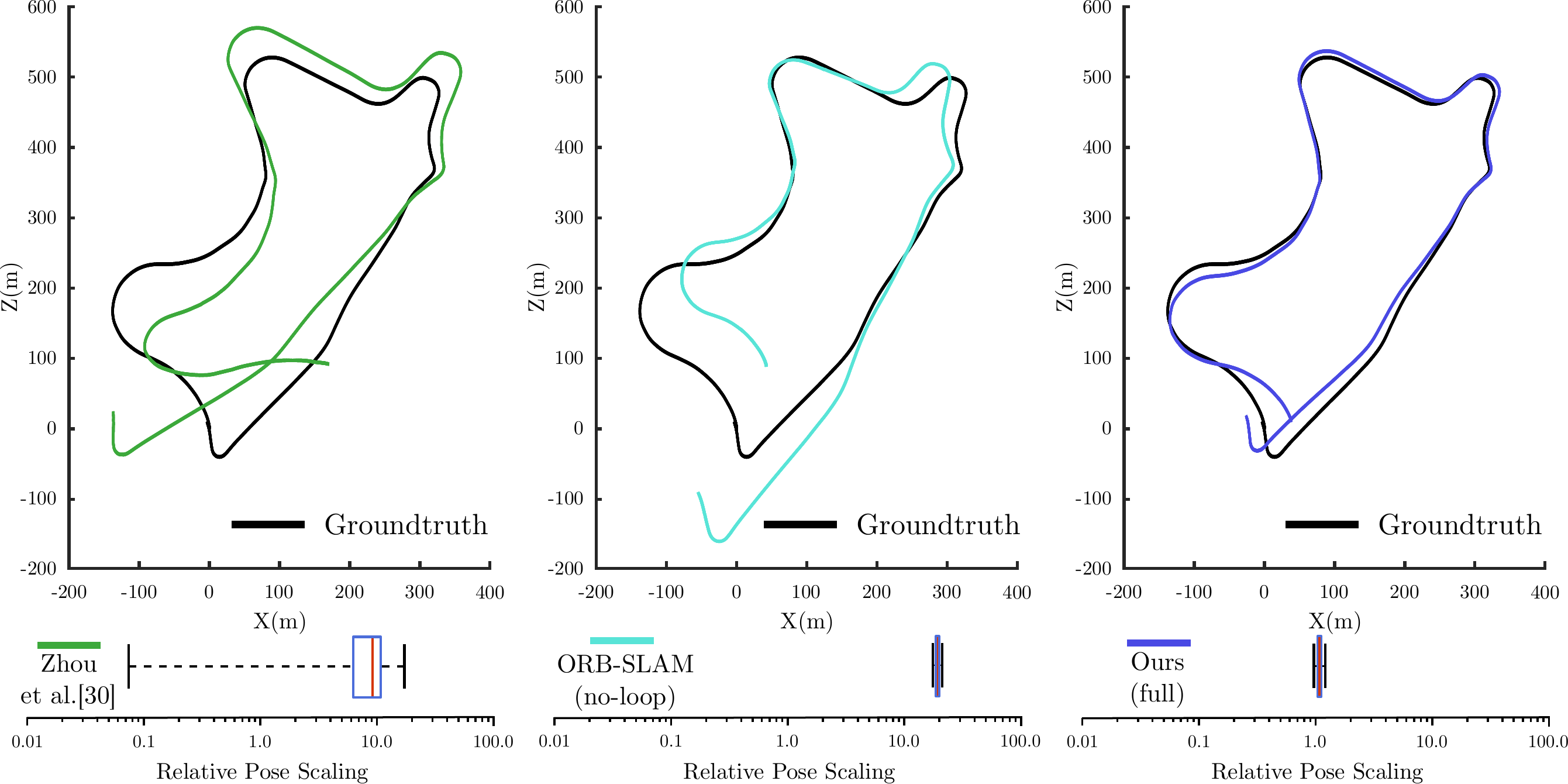}
		\caption{\emph{Top} the scaled and aligned trajectories for Zhou \textit{et al.}\cite{zhou2017unsupervised}(SfM-Learner), ORB-SLAM2 \cite{mur2015orb} (without loop-closure enabled) and Ours respectively. \emph{Bottom} box-plots of the relative pose scaling required to bring the predicted translation to the same magnitude as the ground-truth pose}
		\label{fig:pose_scale_driftseq09}
	\end{center}
\end{figure}
In Table \ref{tab:kitti_pose} we show the most comparable performance of our approach to state-of-the-art SLAM systems. We demonstrate a noticeable improvement over SfM-Learner on both sequences in all metrics. We evaluate SfM-Learner on its frame-to-frame tracking performance for adjacent frames (\emph{SFM-Learner(1)}) and separations of 5 frames (\emph{SFM-Learner(5)}), as they train their approach to estimate this size frame gap. Even with the massive reduction in accumulation error expected by taking larger frame gaps (demonstrated in reduced ATE) our system still produces more accurate pose estimates.
\begin{figure}[h!]
	\begin{center}
    	\includegraphics[width=.9\linewidth]{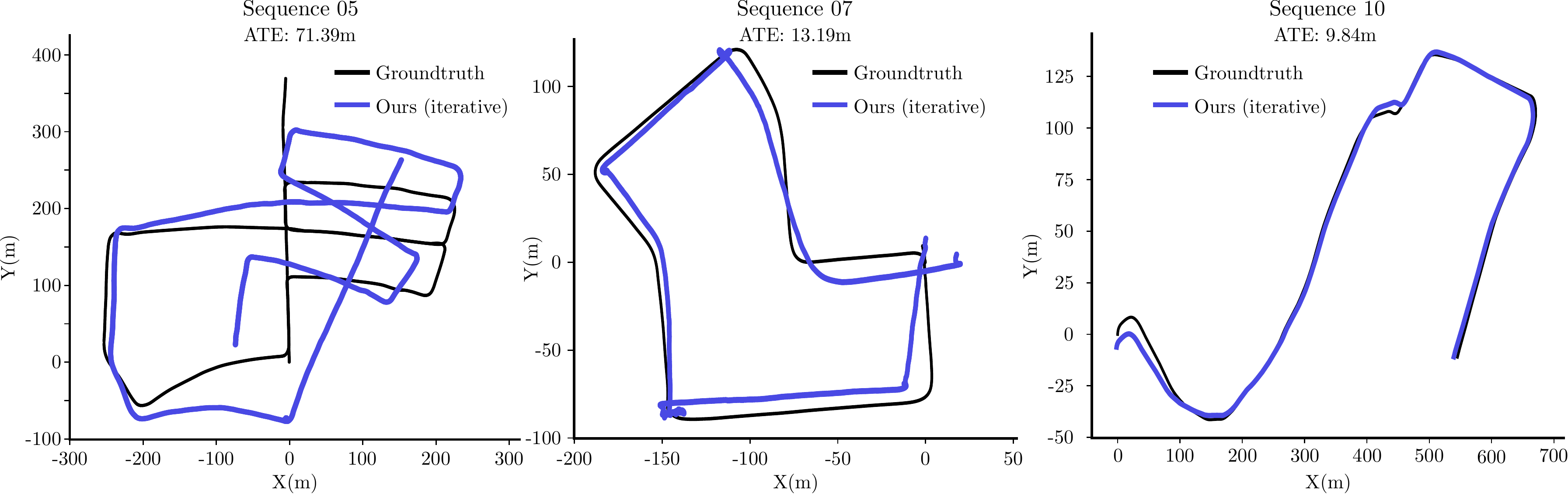}
		\caption{Resulting trajectories  using our iterative approach on 3 additional sequences of KITTI \cite{Geiger2013IJRR}. Sequence 05 shows a failure for our approach, where accumulated drift causes the trajectory to not be well aligned. These sequences are not used for training}
		\label{fig:kitti_sequences}
	\end{center}
\end{figure}
We show the resulting scaled trajectories of sequence 09 in Figure \ref{fig:pose_scale_driftseq09}, as well as the relative scaling of each trajectories poses in a box-plot. The spread of scales present for SFM-Learner indicates scale is essentially ignored by their system, with scale drifts ranging across a full log scale, while ORB-SLAM and our approach are barely visible at this scale. Another thing to note is that our scale is centered around 1.0, as we estimate scale directly by estimating metric depths. This seems to provide a strong benefit in terms of reducing scale-drift and we believe makes our system more usable in practice.

In Table \ref{tab:rgbd_pose} we show a significant improvement in performance against existing machine learning approaches across several sequences from the RGB-D dataset\cite{sturm12iros}. We evaluate against DeMoN\cite{UZUMIDB17} in two ways, frame-to-frame (\emph{DeMoN(1)}) and we again try to provide the same advantage to DeMoN as SfM-Learner by using wider baselines, which they claim improves their depth estimations\cite{UZUMIDB17}, using a frame gap of 10 (\emph{DeMoN(10)}). It can be observed that even with the massive reduction in accumulation error over our frame-to-frame approach, we still manage to significantly out-perform their approach in ATE, even surpassing LSD-SLAM on the sequence \emph{fr1-xyz}. ORB-SLAM is still the clear winner, as they massively benefit from the ability to perform local bundle-adjustments on the sequences used, which are short trajectories of small scenes. We include an example of a frame from the sequence \emph{fr3-walk-xyz} in Figure \ref{fig:flow_confidence}, which shows this scene is not static, but our system has the ability to deal with this through the flow confidence estimates, discussed in Section \ref{subsec:ablation_studies}
\begin{table}[h!]
\centering
\caption{Performance of several approaches evaluated on two sequences of the KITTI dataset \cite{Geiger2013IJRR}. SfM-Learner(1) and SfM-Learner(5) indicates the different frame gaps used to construct the trajectories. The results are separated by SLAM and machine learning approaches. We highlight the strongest results in bold for each type of approach.}
\begin{tabular}{cY{1.35cm}Y{1.35cm}Y{1.35cm}Y{1.35cm}Y{1.35cm}Y{1.35cm}}
    \toprule
	Sequence & \multicolumn{3}{c}{09} & \multicolumn{3}{c}{10} \\ \midrule
    \multirow{1}{*}{Method} & ATE(m) & RPE(m) & RPE(\degree) & ATE(m) & RPE(m) & RPE(\degree) \\ \midrule
    ORB-SLAM(no-loop)\cite{mur2015orb} & 57.57 & \textbf{0.040} & 0.103 & 8.090 & 0.033 & 0.105 \\ 
    ORB-SLAM(full)\cite{mur2015orb} & \textbf{9.104} & 0.056 & \textbf{0.084} & \textbf{7.349 }& \textbf{0.031} & \textbf{0.100} \\ \midrule
    SfM-learner(5)\cite{zhou2017unsupervised} & 58.31 & 0.077 & 0.803 & 31.75 & 0.069 & 1.242 \\ 
    SfM-learner(1)\cite{zhou2017unsupervised} & 81.09 & 0.050 & 0.976 & 75.89 & 0.045 & 1.599 \\
    Ours(fully connected) & 41.50 & 0.087 & 0.387 & 29.29 & 0.081 & 0.486 \\ 
    Ours(full) & \textbf{16.55} & \textbf{0.047} & \textbf{0.128} & \textbf{9.846} & \textbf{0.039} & \textbf{0.138} \\
\end{tabular}
\label{tab:kitti_pose}
\end{table}
\begin{table}[h!]
\centering
\caption{Performance of pose estimation on several sequences from the RGB-D dataset \cite{sturm12iros}. DeMoN(1) and DeMoN(10) indicates the trajectories were constructed with a frame gap of 1 and 10 respectively. Both \cite{mur2015orb} and \cite{engel2014lsd} fail to track on \emph{fr2-360-hs}. The results are separated by SLAM and machine learning approaches. We highlight the strongest results in bold for each type of approach.}
\begin{tabular}{cY{1cm}Y{1cm}Y{1cm}Y{1cm}Y{1cm}Y{1cm}Y{1cm}Y{1cm}Y{1cm}}
    \toprule
	Sequence & \multicolumn{3}{c}{fr1-xyz} & \multicolumn{3}{c}{fr2-360-hs} & \multicolumn{3}{c}{fr3-walk-xyz} \\ \midrule
    \multirow{2}{*}{Method} & ATE & RPE & RPE & ATE & RPE & RPE & ATE & RPE & RPE \\ 
    & (m) & (m) & (\degree) & (m) & (m) & (\degree) & (m) & (m) & (\degree) \\\midrule
    LSD-SLAM\cite{engel2014lsd} & 0.090 & - & - & - & - & - & 0.124 & - & -\\
    ORB-SLAM\cite{mur2015orb} & \textbf{0.009} & \textbf{0.007} & \textbf{0.645} & - & - & - & \textbf{0.012} & \textbf{0.013} & \textbf{0.694}\\ \midrule
       DeMoN(10)\cite{UZUMIDB17} & 0.178 & \textbf{0.021} & \textbf{1.193} & 0.601 & 0.035 & 2.243 & 0.265 & 0.049 & 1.447\\
    DeMoN(1)\cite{UZUMIDB17} & 0.183 & 0.037 & 3.612 & 0.669 & 0.032 & 3.233 & 0.279 & 0.040 & 3.174\\

      Ours(fully connected) & 0.169 & 0.028 & 1.887 & {0.883} & {0.030} & {1.799} &0.268 & 0.044& 1.698\\
    Ours(iterative) & \textbf{ 0.071} & 0.024 & 1.237 & \textbf{0.461} & \textbf{0.020} & \textbf{0.736} & \textbf{0.240} & \textbf{0.026} & \textbf{0.811}
\end{tabular}
\label{tab:rgbd_pose}
\end{table}

\subsection{Ablation Experiments}
\label{subsec:ablation_studies}
In order to examine the contribution of using each component of our pose estimation network, we compare the pose estimates under various configurations on sequences 09 and 10 of the KITTI odometry dataset\cite{Geiger2013IJRR}, summarised in Table \ref{tab:kitti_confidence}. We examine the relative improvement of iterating on our pose estimation till convergence, against a single weighted-least-squares iteration, which demonstrates iterating has a significantly positive effect. We demonstrate the improved utility of our flows by replacing our flow estimates with other state-of-the-art flow estimation methods from \cite{ilg2017flownet} and \cite{revaud2015epicflow} in our pose estimation pipeline, and consistently demonstrate an improvement using our approach. We show the result of optimising with and without our estimated confidences, demonstrating quantitatively how important they are to pose estimation accuracy, with significant reductions across all metrics.
\begin{table}[h!]
\centering
\caption{Results of pose estimation on KITTI\cite{Geiger2013IJRR} with various components of the network removed or replaced. We highlight the strongest results in bold.}
  \begin{tabular}{cY{1.4cm}Y{1.4cm}Y{1.4cm}Y{1.4cm}Y{1.4cm}Y{1.4cm}}
      \toprule
      Sequence & \multicolumn{3}{c}{09} & \multicolumn{3}{c}{10} \\ \toprule
      \multirow{1}{*}{Method} & ATE(m) & RPE(m) & RPE(\degree) & ATE(m) & RPE(m) & RPE(\degree) \\ \midrule
      Ours(noconf) & 53.40 & 0.356 & 0.931 & 58.50 & 0.308 & 1.058 \\
      Ours(noconf,iterative) & 33.18 & 0.248 & 0.421 & 35.87 & 0.280 & 0.803 \\
      Flownet2.0\cite{ilg2017flownet} & 29.64 & 0.349 & 0.838 & 51.90 & 0.222 & 0.954 \\
      Flownet2.0(iterative)\cite{ilg2017flownet} & 24.61 & 0.185 & 0.400 & 22.61 & 0.142 & 0.484 \\
      EpicFlow\cite{revaud2015epicflow} & 119.0 & 0.566 & 0.931 & 20.98 & 0.199 & 0.853 \\
      EpicFlow(iterative)\cite{revaud2015epicflow} & 59.79 & 0.379 & 0.459 & 14.80 & 0.154 & 0.581 \\
      Ours(full-single iteration) & 31.20 & 0.089 & 0.324 & 24.10 & 0.095 & 0.389 \\
      Ours(full-til convergence) & \textbf{16.55} &\textbf{ 0.047} & \textbf{0.128 }& \textbf{9.846} & \textbf{0.039} & \textbf{0.138}
  \end{tabular} 
\label{tab:kitti_confidence}
\end{table}

\begin{figure}[h!]
	\begin{center}
    	\includegraphics[width=\linewidth]{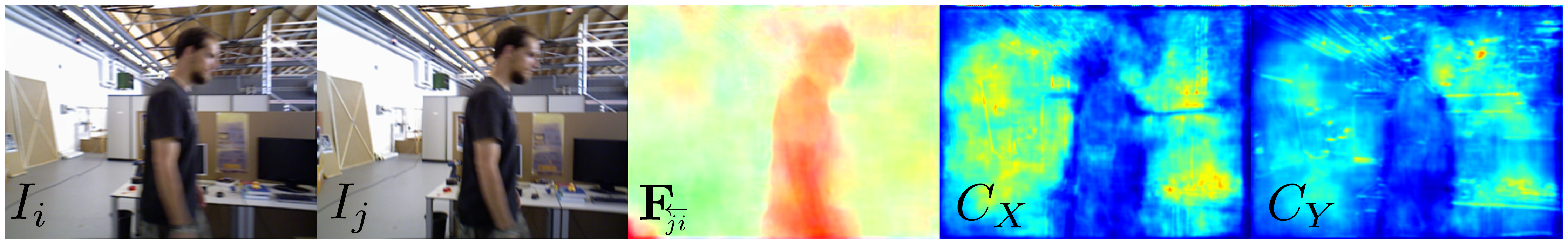}
		\caption{For a frame pair ($I_i$ and $I_j$) from the sequence \emph{fr3-walk-xyz} , $\mathbf{F}_{\protect \overleftarrow{ji}}$ is the estimated optical flow from $I_i$ to $I_j$, and $C_x$ and $C_y$ are the estimated flow confidences in the $x$ and $y$ direction respectively}
		\label{fig:flow_confidence}
	\end{center}
\end{figure}

We also demonstrate qualitatively one of the ways in which estimating confidence improves our pose estimation in Figure \ref{fig:flow_confidence}. This shows that our system has learned the confidence on moving objects is lower than its surroundings and the confidences of edges are higher, helping our system focus on salient information during optimisation in an approach similar to \cite{engel2014lsd}.

\section{Conclusion and Further Work}

We present the first piece of work that performs least squares based pose estimation inside a neural network. Instead of replacing every component of the SLAM pipeline with CNNs, we argue it's better to use CNNs for tasks that greatly benefit from feature extraction (depth and flow prediction) and use geometry for tasks its proven to work well (motion estimation given the depths and flow). Our formulation is fully differentiable and is trained end-to-end. We achieve state-of-the-art performances on single image depth prediction for both NYUv2 \cite{Silberman} and KITTI \cite{Geiger2013IJRR} datasets. We demonstrate both qualitatively and quantitatively that our system is capable of producing better visual odometry that considerably reduces scale-drift by predicting metric depths. 

\clearpage
\section*{Supplementary Material}
\setcounter{section}{0}
\renewcommand{\thesection}{\Roman{section}} 
\renewcommand{\thesubsection}{\thesection.\Roman{subsection}}
\section{Dataset Evaluation Analysis}

In this section we evaluate and analyse the relative performance on each dataset as well as correlations in the dataset and how they relate to overall performance.  

\subsection{NYUv2\cite{Silberman}}

\begin{figure}[h!]
	\begin{center}
    	\includegraphics[width=\linewidth]{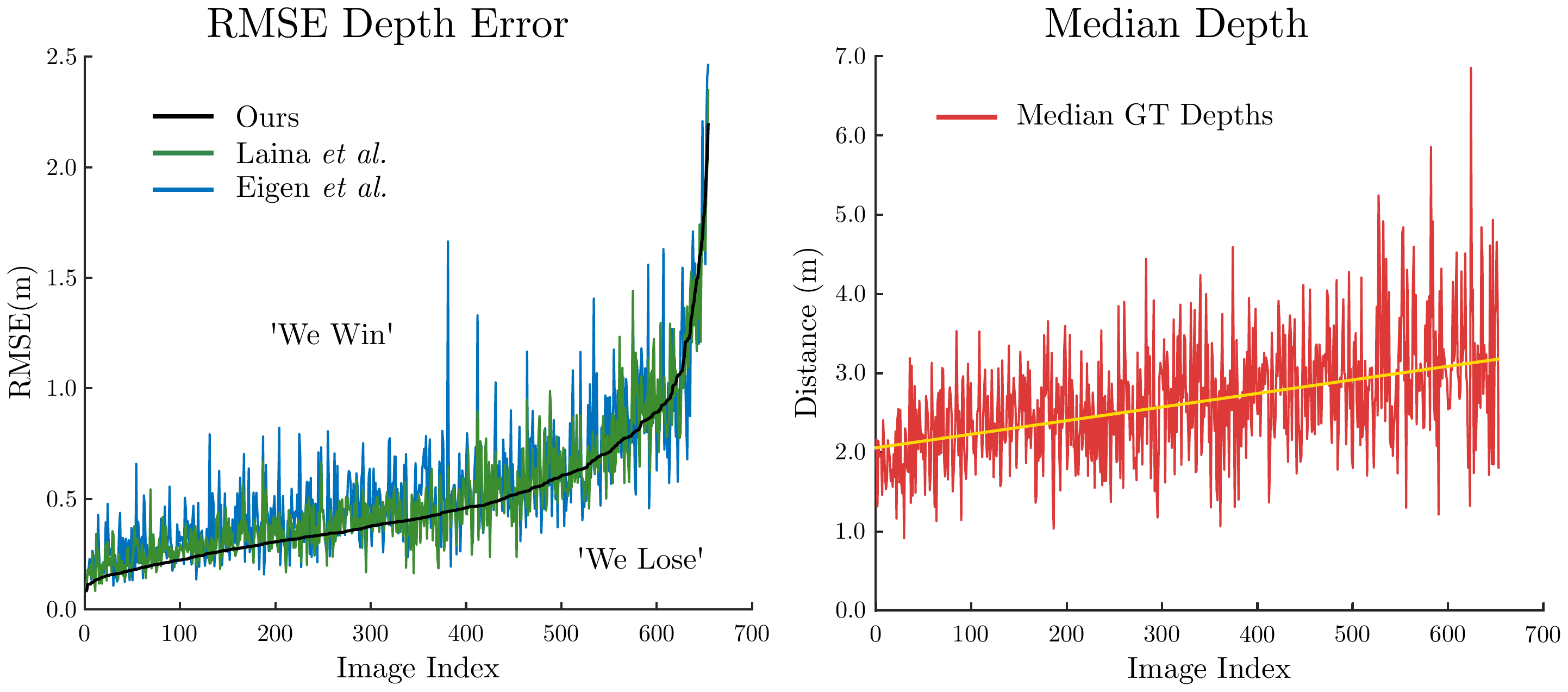}
		\caption{\emph{Left:} The RMSE error on each image of the test set,  sorted by our performance on the NYUv2 dataset. We include two competing approaches, as well as marking which side of the line indicates we are better (\emph{`We Win'}) and which we are worse (\emph{`We Lose'}). \emph{Right:} The median ground-truth depth of each image in the test set also sorted by our RMSE performance. We include an approximate trend-line to show the relationship between depth and RMSE in our system}
		\label{fig:dataset_sorted_performance}
	\end{center}
\end{figure}

\begin{figure}[h!]
	\begin{center}
    	\includegraphics[width=\linewidth]{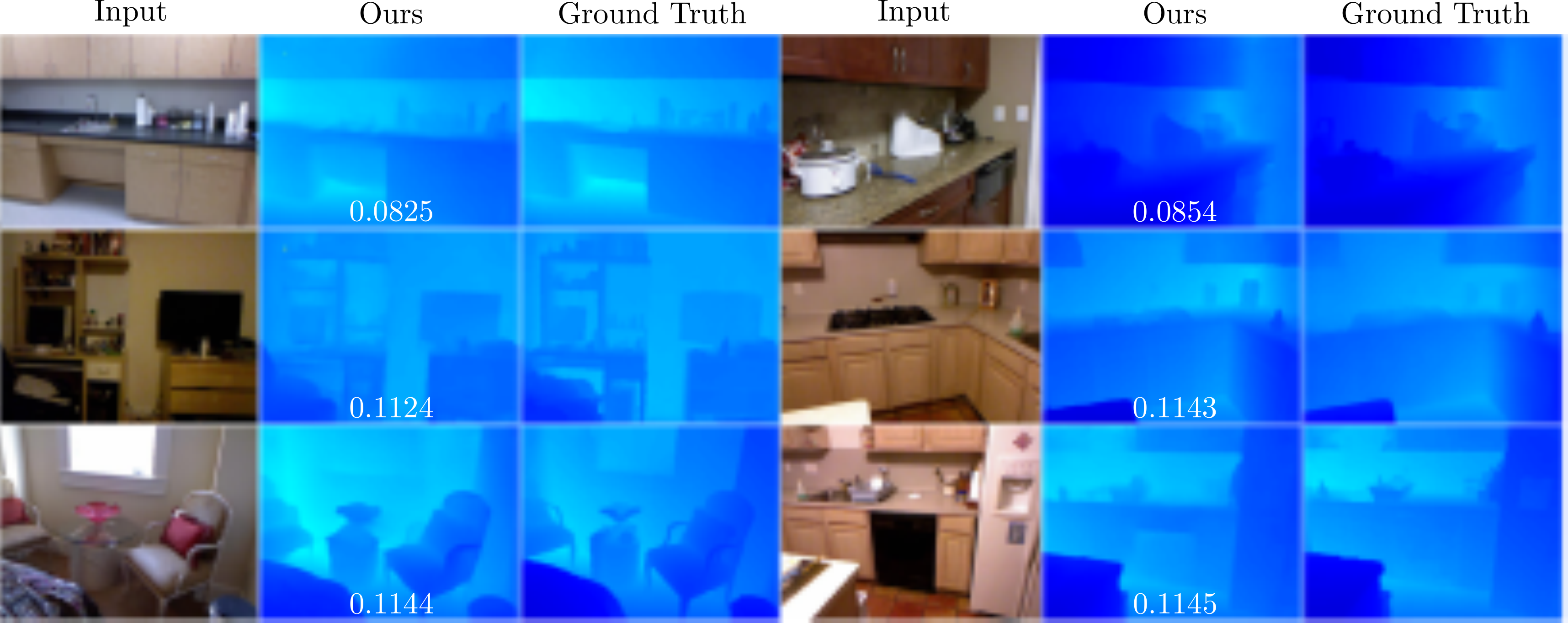}
		\caption{The 6 highest performing images from the NYUv2 \cite{Silberman} testset, based on RMSE error. All images are of varying scenes, but contain lower median depth values on average.}
		\label{fig:nyu_best}
	\end{center}
    \vspace*{-3em}
\end{figure}

\begin{figure}[h!]
	\begin{center}
    	\includegraphics[width=\linewidth]{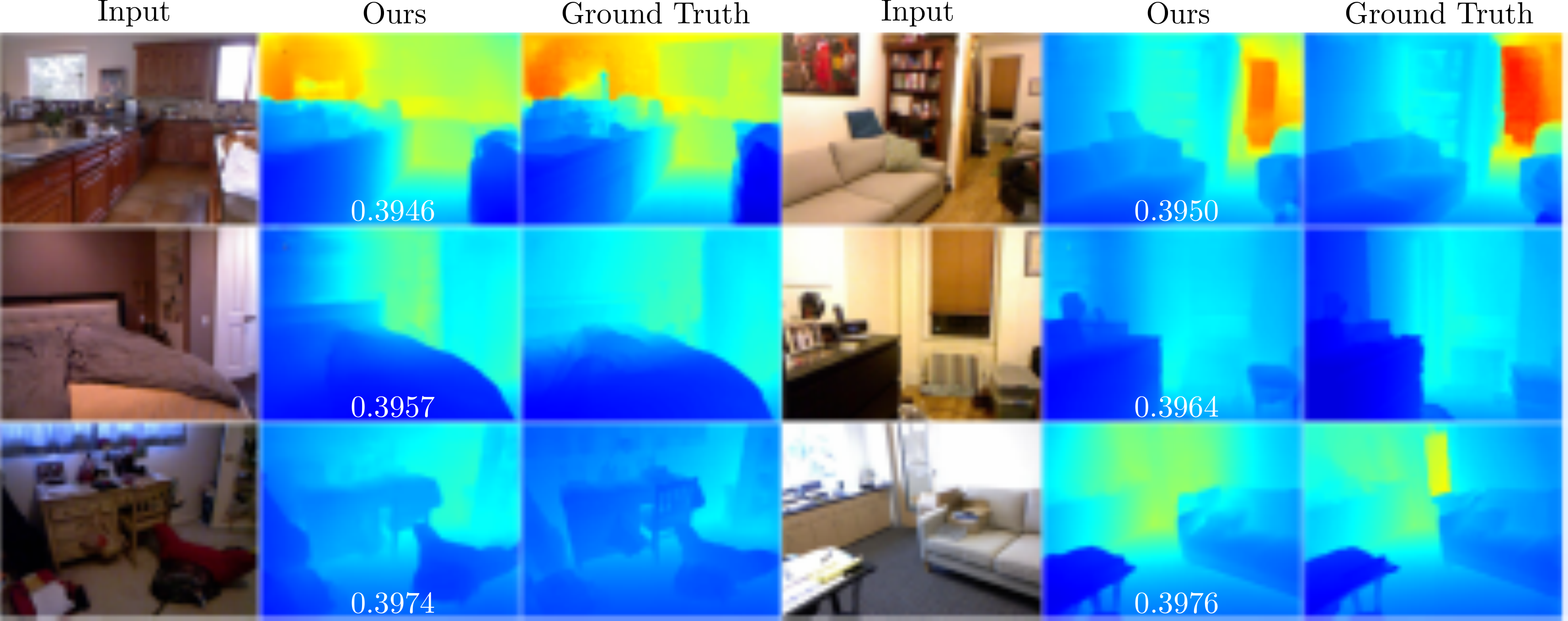}
		\caption{The middle 6 images from the NYUv2 \cite{Silberman} testset, based on RMSE error. All images have RMSE values individually lower than the full testset (0.480m), indicating a small number of outliers, which is apparent in Figure \ref{fig:dataset_sorted_performance}}
		\label{fig:nyu_middle}
	\end{center}
    \vspace*{-1em}
\end{figure}

The dataset NYUv2\cite{Silberman} has been a popular benchmark for indoor depth estimation and semantic segmentation since the work of Eigen \emph{et al.}\cite{Eigen}. We provide several qualitative and quantitative results from the evaluation of our approach in Figures \ref{fig:nyu_best}, \ref{fig:nyu_middle} and \ref{fig:nyu_worst}. This shows our strongest, median and worst performing images, as well as each predictions RMSE error in meters. This reveals two insights about our system's performance, and that is we perform stronger on images with closer median depths and that our largest errors occur when we incorrectly estimate the overall scale of the scene. The relationship to median depth is evident in Figure \ref{fig:dataset_sorted_performance}, where the RMSE is strongly correlated to the median scene depth. We also observe a similarly strong correlation in the performance of all three approaches, although our approach is overwhelmingly out performing the competitors.

\begin{figure}[h!]
	\begin{center}
    	\includegraphics[width=\linewidth]{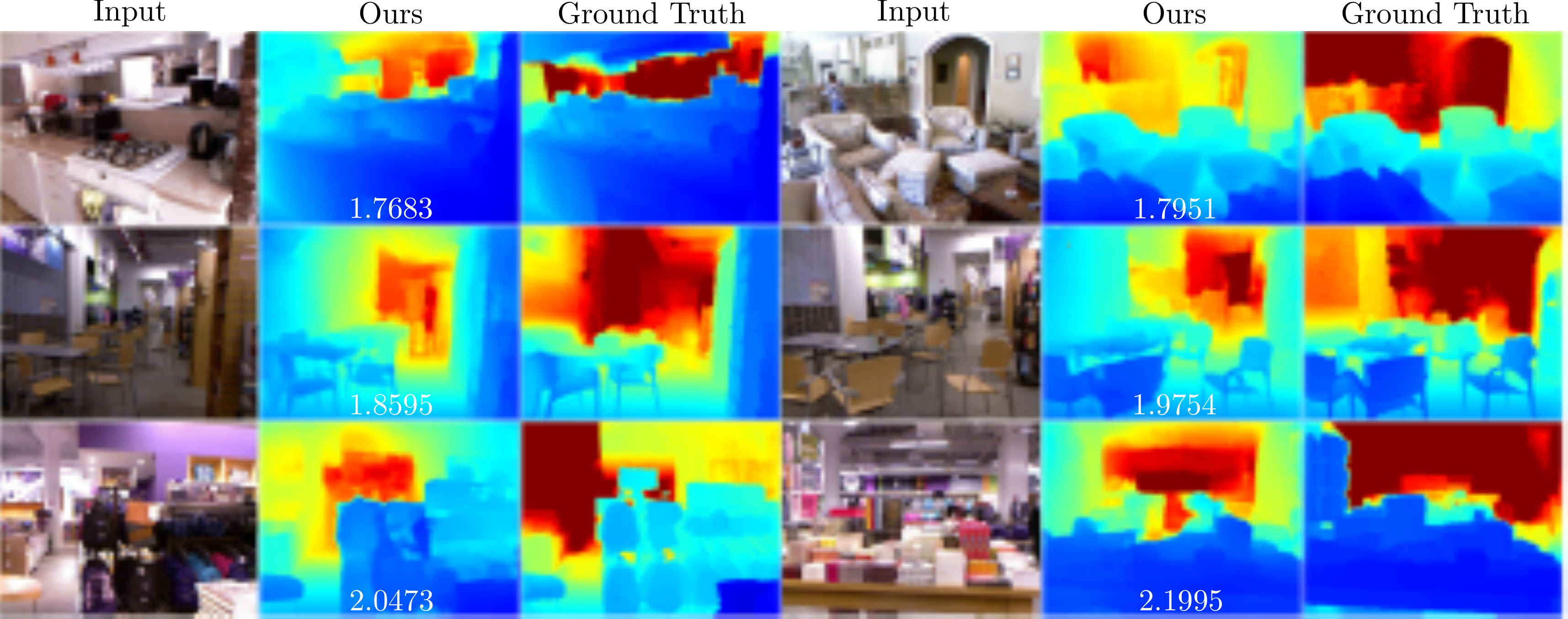}
		\caption{The 6 lowest performing images from the NYUv2 \cite{Silberman} testset, based on RMSE error. In general these images contain higher median depth values, but the way in which our network gets it wrong appears to be in estimating the overall scene scale. This quantity is challenging to estimate, and we can observe qualitatively the system still produces believable relative depth estimates}
		\label{fig:nyu_worst}
	\end{center}
    \vspace*{-3em}
\end{figure}

What conclusions can we draw from these results? Well this is a rather clear result of the choice of error metric in ranking the results. In this case as we rank by the RMSE, we would expect higher depths to be the images with the largest error, as only either very large predictions or very large ground truth values can generate large RMSE values. This also indicates that our network tends to behave conservatively, estimating the scene is closer on average rather than further. This is probably a direct result of the depth value distribution in the training set, potentially biasing the depths towards the lower end.

\clearpage
\subsection{KITTI \cite{Geiger2013IJRR}}
\begin{figure}[h!]
	\begin{center}
    	\includegraphics[width=\linewidth]{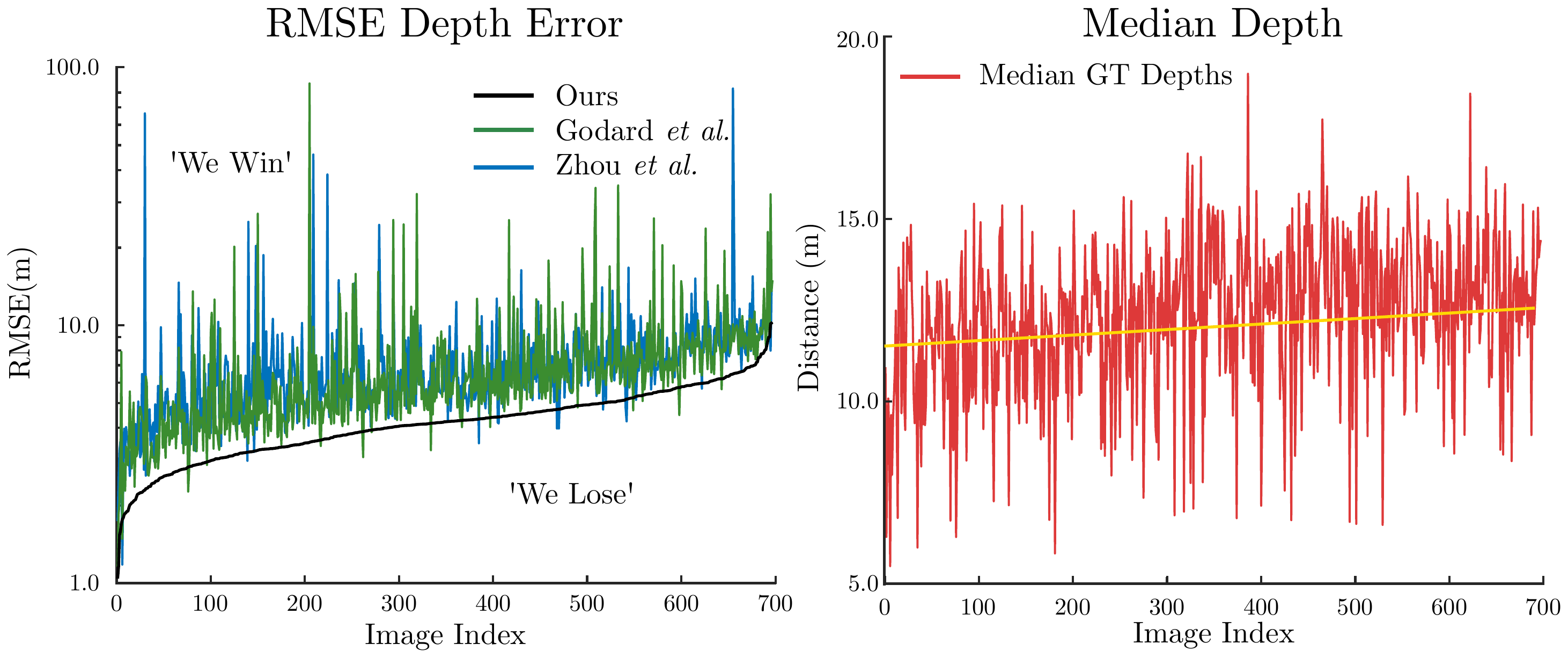}
		\caption{\emph{Left:} The RMSE error on each image of the test set,  sorted by our performance on the KITTI dataset. We include two competing approaches, as well as marking which side of the line indicates we are better(\emph{`We Win'}) and which we are worse (\emph{`We Lose'}). \emph{Right:} The median ground-truth depth of each image in the test set also sorted by our RMSE performance. We include an approximate trend-line to show the relationship between depth and RMSE in our system}
		\label{fig:dataset_sorted_performance_kitti}
	\end{center}
    \vspace*{-3em}
\end{figure}

\begin{figure}[h!]
	\begin{center}
    	\includegraphics[width=\linewidth]{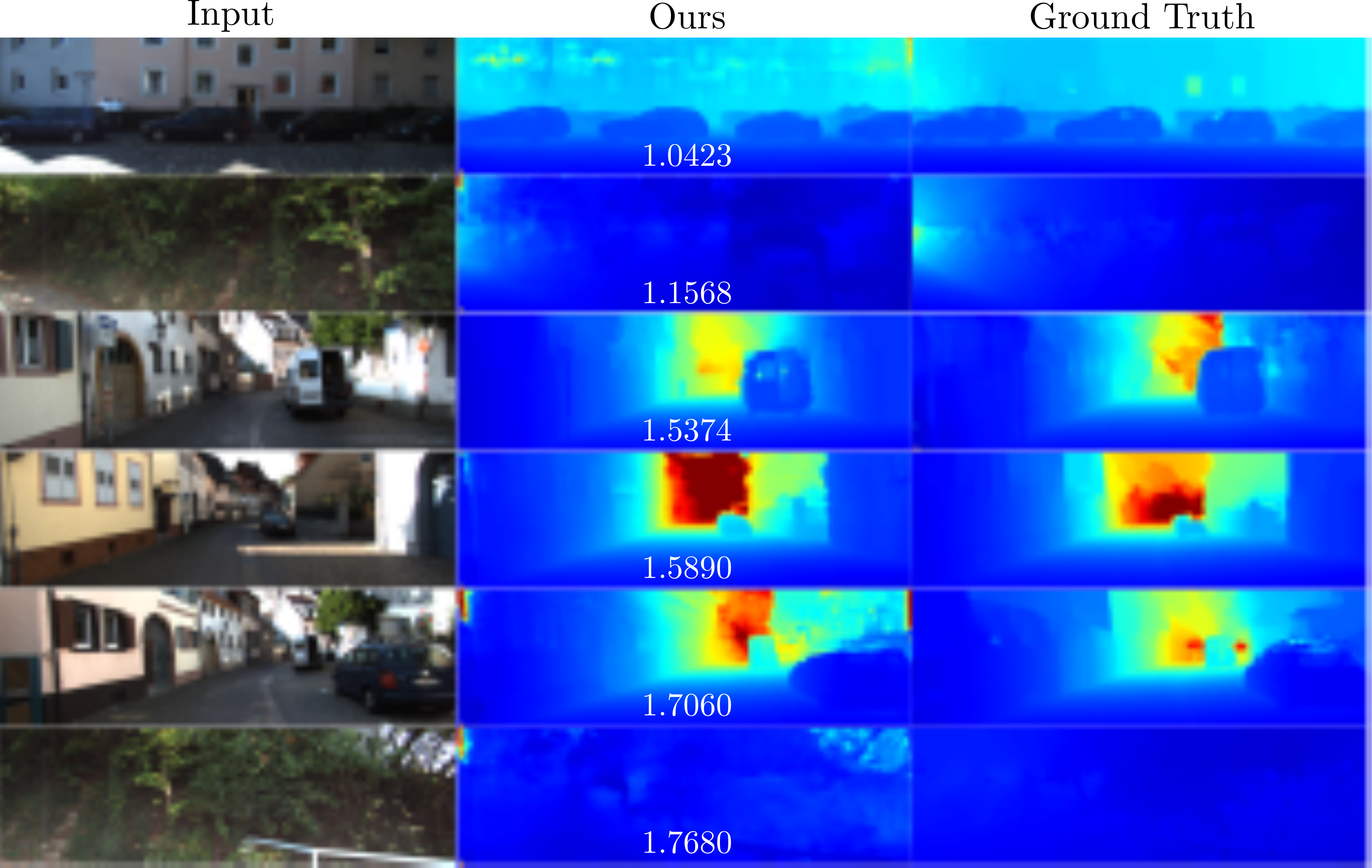}
		\caption{The highest performing 6 images from the KITTI \cite{Geiger2013IJRR} testset, based on RMSE error. Surprisingly not all of these contain a great deal of scale context, in particular rows 2 and 6, where they face a dirt ramp, which is atypical of the predominantly road facing dataset. This indicates strongly that the approach is genuinely learning about the geometry of the scenes}
		\label{fig:kitti_best}
	\end{center}
    \vspace*{-3em}
\end{figure}

\begin{figure}[h!]
	\begin{center}
    	\includegraphics[width=0.96\linewidth]{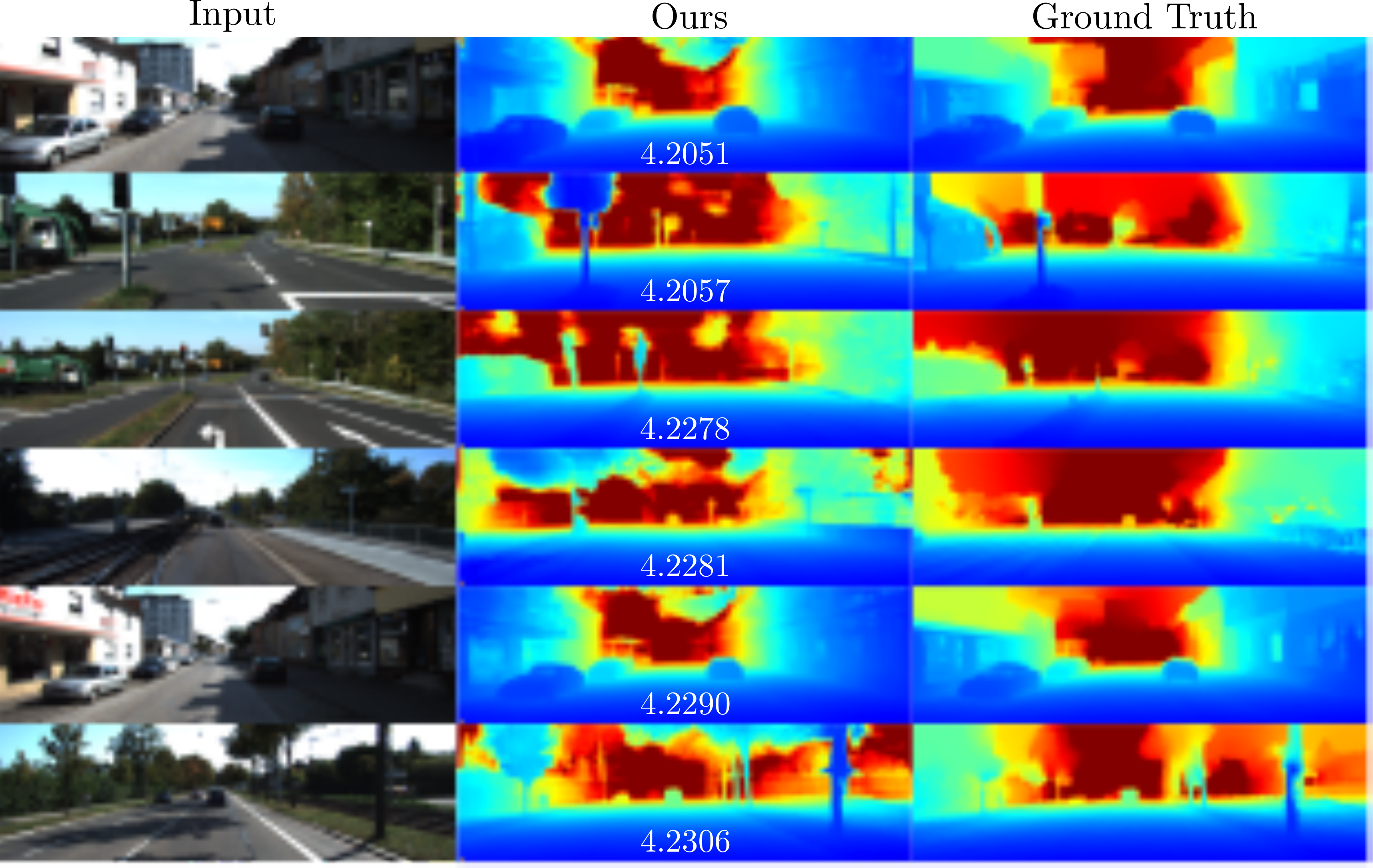}
		\caption{The middle 6 performing images from the KITTI \cite{Geiger2013IJRR} testset, based on RMSE error. The RMSE values are hovering around the value achieved for the dataset and represent the typical performance. Note the systems ability to estimate depth in the top half of the scene, which never receives a ground truth training signal, as the LIDAR only scans below the horizon line}
		\label{fig:kitti_middle}
	\end{center}
\end{figure}

\begin{figure}[h!]
\vspace*{-2em}
	\begin{center}
    	\includegraphics[width=0.96\linewidth]{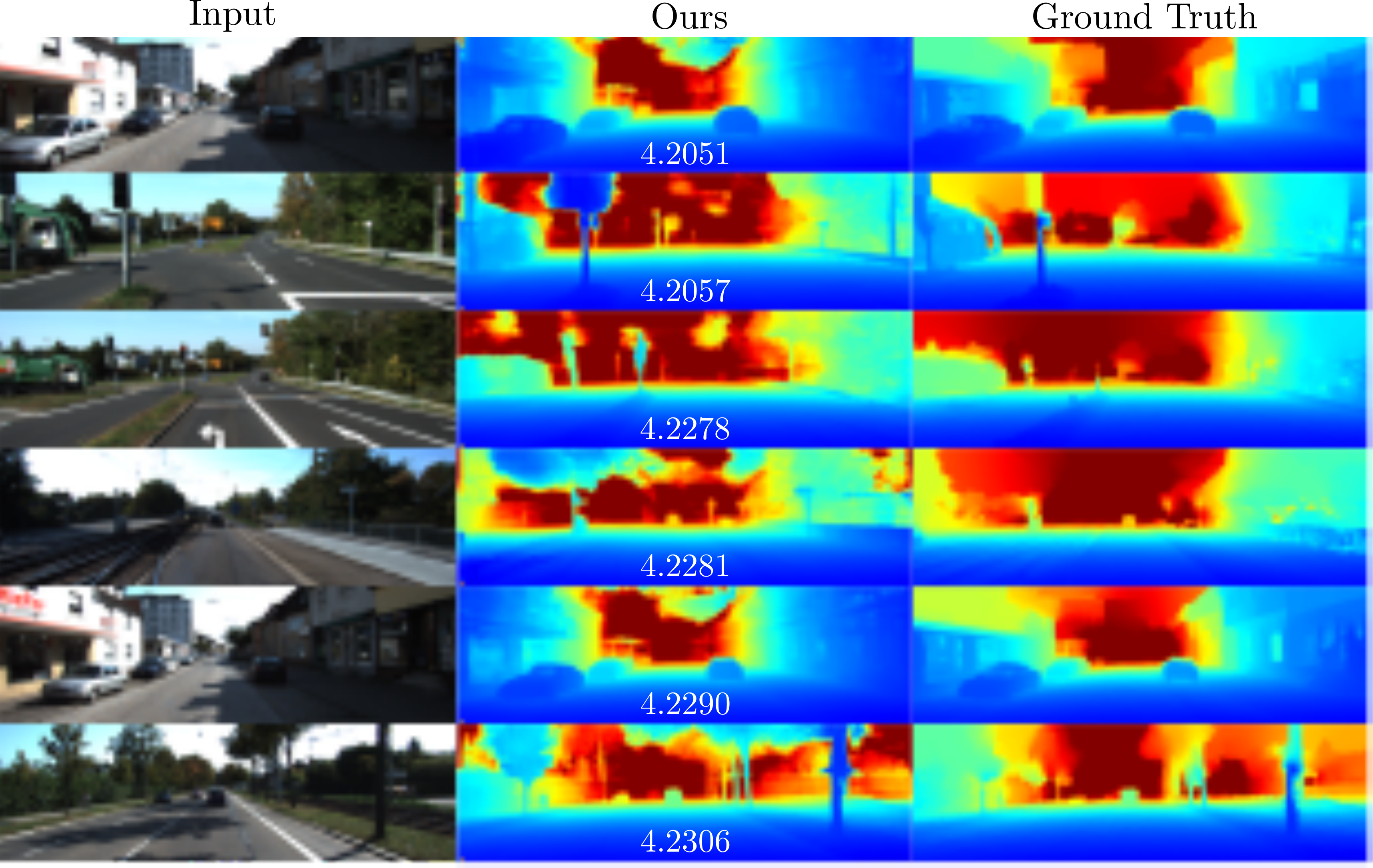}
		\caption{The lowest 6 performing images from the KITTI \cite{Geiger2013IJRR} testset, based on RMSE error. Although the RMSE of each of these images is comparitively high, the depth predictions produced are convincing qualitatively}
		\label{fig:kitti_worst}
	\end{center}
    \vspace*{-3em}
\end{figure}

Our most impressive performance is perhaps on the KITTI benchmark dataset \cite{Geiger2013IJRR}. Where as shown in Figure \ref{fig:dataset_sorted_performance_kitti} (\emph{left}), we consistently out-perform the competing approaches on almost all test images. The scale of the depth error had to be changed to $\log_{10}$ in order to capture the full range of errors. This could be because the competing approaches estimate inverse  depth/disparity and invert the predicted values to compute their loss function. This can lead to unstable performance on large distances, due to the non-linearity of this section, as opposed to our approach which is linear to all depths.  

Again we include the analysis of the median depths sorted against the RMSE error in Figure \ref{fig:dataset_sorted_performance_kitti} (\emph{right}), as we did for NYUv2. In this case the relationship between error and depth is largely reduced, this is most likely due to the nature of the dataset which contains a very similar spread of data for most images in the training set, as they film similar scenarios. However the relationship is still visible in Figure \ref{fig:kitti_best}, where these scenes contain comparitively low depth values, indicating again our system behaves conservatively in estimating depths.

\clearpage

\subsection{Comparison using the architecture of Kuznietsov \cite{kuznietsov2017semi}}

We replaced the architecture of our depth estimation network using that of Kuznietsov \emph{et al.} \cite{kuznietsov2017semi}. As it can be seen below by using the full training loss we are able to improve the accuracy of the depth estimation results indicating the generality of the approach. 

\begin{table}
	\centering
    \vspace*{-0.5em}
        \begin{tabular}{Y{0.5cm}lY{1.5cm}Y{1.5cm}Y{1.5cm}Y{1.5cm}Y{1.5cm}Y{1.5cm}}
        \toprule
        & \multirow{2}{*}{Dataset} & \multicolumn{2}{c}{\textbf{NYU}\cite{Silberman}} & \multicolumn{2}{c}{\textbf{TUM}\cite{sturm12iros}} &\multicolumn{2}{c}{\textbf{KITTI} \cite{Geiger2013IJRR}} \\ 
         & & Baseline & Full & Baseline & Full & Baseline & Full \\
        \midrule
        \multirow{4}{*}{\rotatebox{90}{\normalsize Metric}} & $\text{RMSE}(m)$    & 0.536 & 0.525 & 1.096 & 1.015 & 3.518 & 3.425    \\
         & $<\delta$ (=1.25)  & 82.5 & 82.8 & 79.9 & 81.1 & 87.5 & 89.5   \\
         & $<\delta^2$ & 96.3 & 96.7 & 90.4 & 91.8 & 96.4 & 96.9 \\
         & $<\delta^2$  & 99.2 & 99.3 &  93.8 &94.6 & 98.8 & 98.8\\
        \bottomrule
        \end{tabular}
        
\label{table:kuz_net}
\end{table}

\clearpage
\pagebreak
\section{Pose Trajectories}

\subsection{KITTI Trajectories}

\begin{figure}[h!]
	\begin{center}
    	\includegraphics[width=\linewidth]{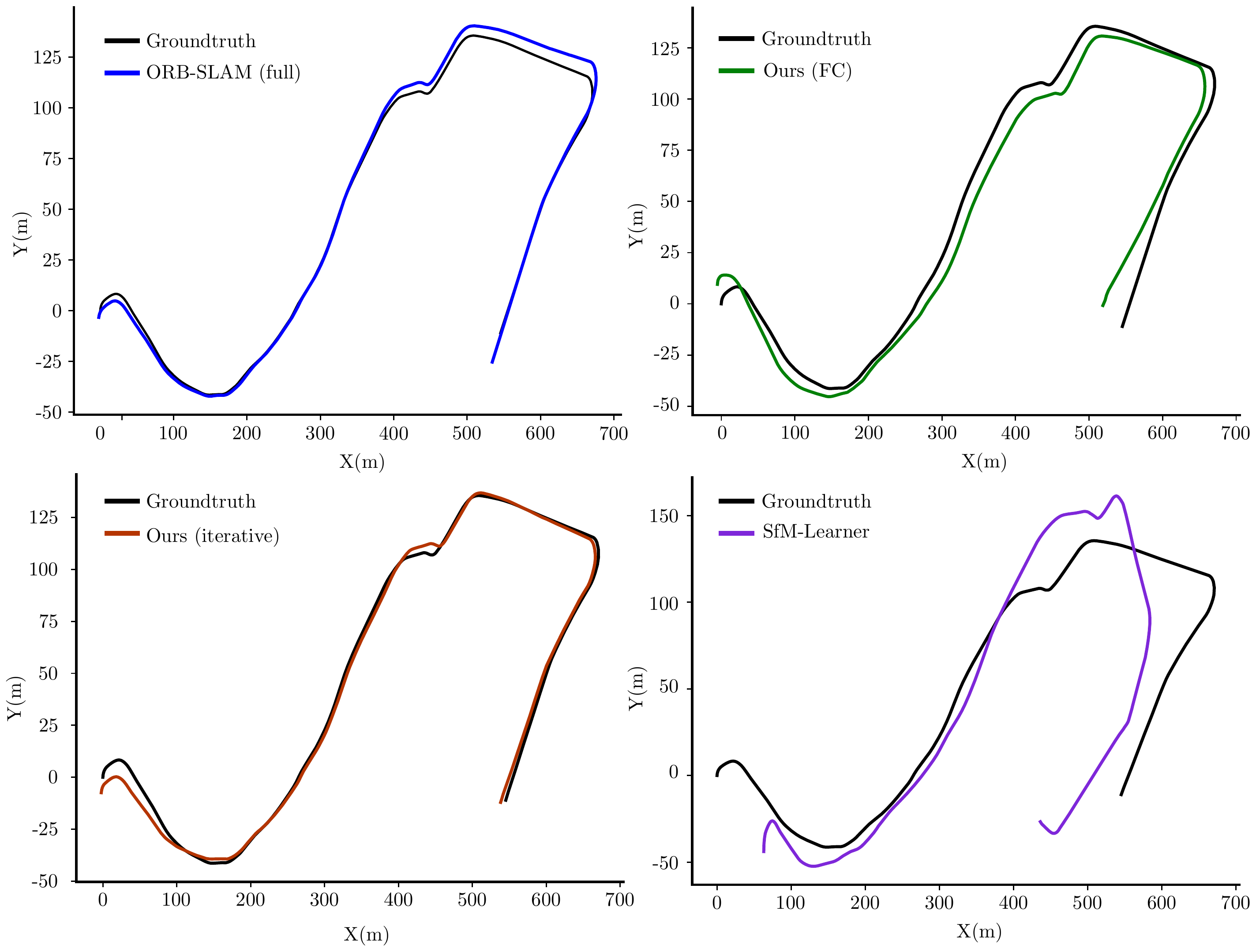}
		\caption{Trajectories of Our method in both configurations, as well as the resulting trajectories of ORB-SLAM(full) \cite{mur2015orb} and SfM-Learner \cite{zhou2017unsupervised}. We demonstrate comparative quality to ORB-SLAM, and significantly out perform SfM-Learner}
		\label{fig:seq10_trajectories}
	\end{center}
    \vspace*{-3em}
\end{figure}

We include the trajectory from sequence 10 of the odometry dataset from KITTI \cite{Geiger2013IJRR}. For the quantitative results please refer to the main paper. The resulting trajectories in Figure \ref{fig:seq10_trajectories}, indicate the comparitively strong performance of our approach, and show that our iterative (\emph{bottom-left}) approach is significantly more accurate than the FC approach (\emph{top-right}). This trajectory contains no loops, and as such could lead to significant scale drift in some SLAM systems, however in this case the frequent local bundle-adjusts performed by ORB-SLAM seem to have helped to maintain the map quality throughout.

\subsection{RGB-D Trajectories}

\begin{figure}[h!]
	\begin{center}
    	\includegraphics[width=\linewidth]{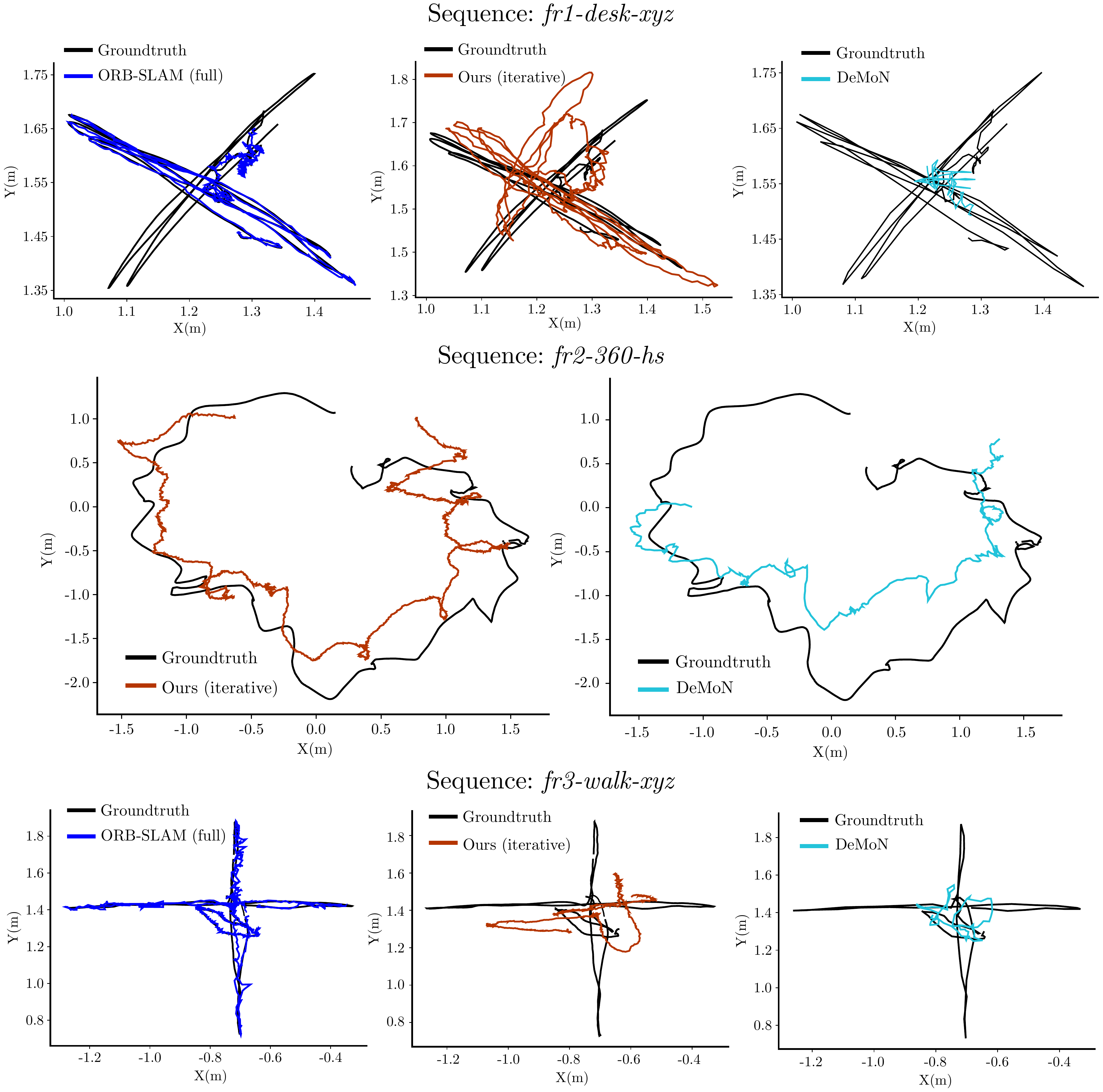}
		\caption{Trajectories of Our method against ORB-SLAM \cite{mur2015orb} and DeMoN(10) \cite{UZUMIDB17}, for the evaluated sequenced from the RGB-D dataset \cite{sturm12iros}. We demonstrate a marked improvement upon DeMoN which, although being given a slight advantage in some respects by widening the baseline and reducing accumulated pose error, still performs poorly. However against ORB-SLAM, both methods come up a little short, as ORB-SLAM is able to perform local bundle-adjustments across multiple keyframes, which greatly reduces the overall error.}
		\label{fig:rgbd_trajectories}
	\end{center}
    \vspace*{-3em}
\end{figure}

We show the estimated aligned trajectories for several sequences from the RGB-D dataset \cite{sturm12iros}, to demonstrate qualitatively performance of our system against previous approaches. We summarise the trajectories in Figure \ref{fig:rgbd_trajectories}, which demonstrates our comparably favourable performance against the approach DeMoN\cite{UZUMIDB17}. This is all despite our method estimating only frame-to-frame relative poses from adjacent frames, while DeMoN(10) is using a larger baseline and thus should estimate a smoother trajectory given the reduction in accumulation error. Ultimately ORB-SLAM \cite{mur2015orb} is still the clear winner, as it uses information from multiple frames, and iteratively aggregates error across short sections. However as our approach is purely VO we were able to get a trajectory for \emph{fr2-360-hs}, which we were unable to for ORB-SLAM due to the challenging nature of the camera motion and rapid lighting changes. 

\subsection{Comparison With CNN-SLAM}
In this table we include a comparison of our approach on the datasets used by CNN-SLAM \cite{tateno2017cnn}. We would like to point out that our method performs competitively despite solely computing sequential frame-to-frame alignments and does not (yet) take advantage of the loop closures and local/global bundle adjustments used by the competing methods.
\begin{table}[h!]
	\centering
    \vspace{-1em}
      \begin{tabular}{lY{2cm}Y{2cm}Y{2cm}Y{2cm}Y{2cm}Y{2cm}}
      \toprule
      \multirow{2}{*}{Method} \hspace{1em} & \multicolumn{3}{c}{Absolute Trajectory Error} \\ \hspace{1em} & $\text{TUM/seq1}$ & $\text{TUM/seq2}$ & $\text{TUM/seq3}$ \\  \midrule
      CNN-SLAM  & \textbf{0.542} & \textbf{0.243} & 0.214\\
      LSD-SLAM & 1.826 & 0.436 & 0.937  \\
      ORB-SLAM & 1.206 & 0.495 & 0.733 \\
      Ours (fc) & 1.043 & 0.672& 0.186 \\
      Ours (full) & 0.799 & 0.587& \textbf{0.157}\\
     \bottomrule
	 \end{tabular}
\label{table:cnn_slam}
\end{table}

\vfill
\pagebreak
\section{Optical Flow}

\begin{figure}[h!]
\vspace*{-2em}
	\begin{center}
    	\includegraphics[width=0.98\linewidth]{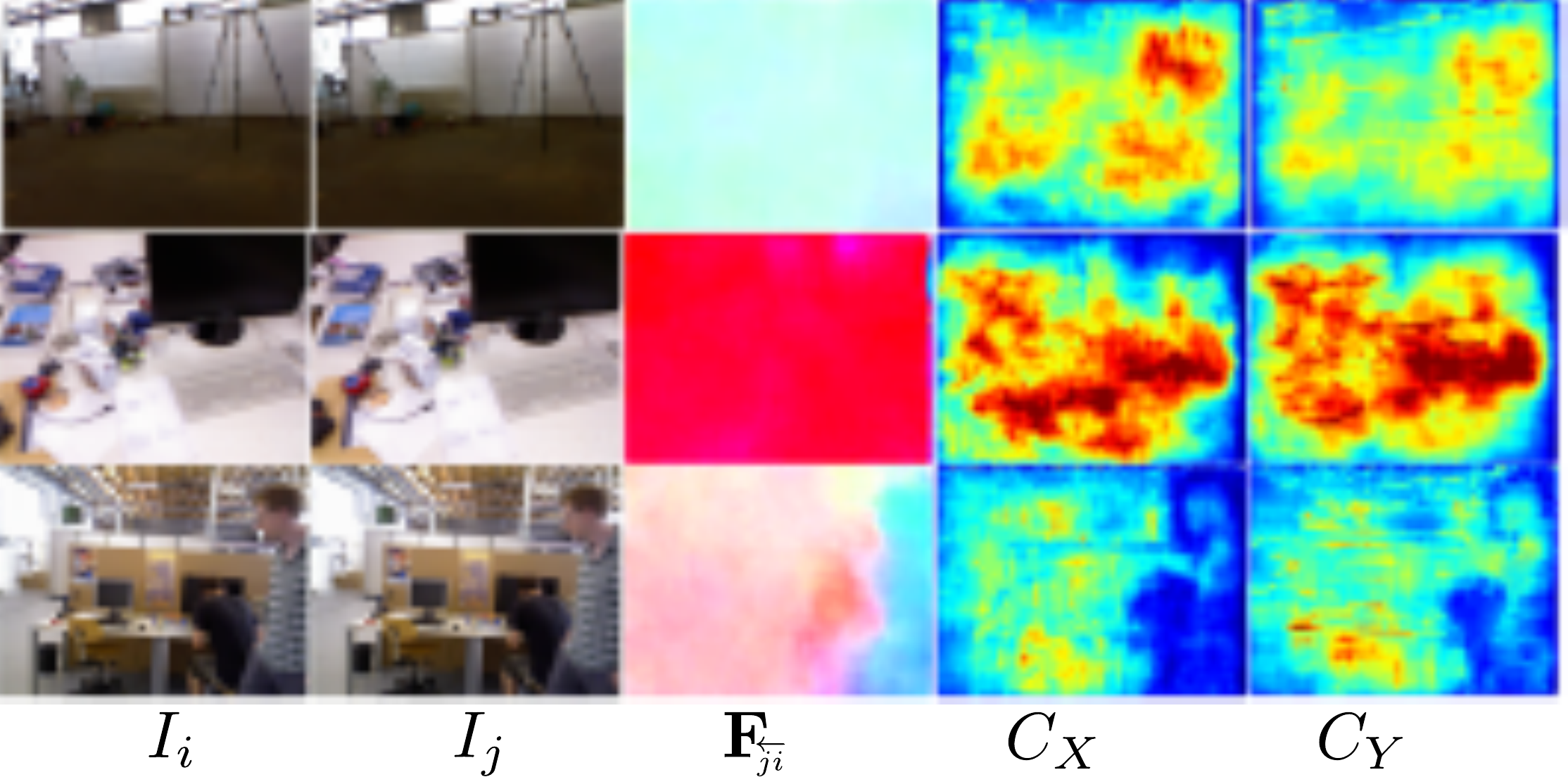}
		\caption{A selection of optical flow predictions made by our framework on the TUM dataset\cite{sturm12iros}. For a frame pair ($I_i$ and $I_j$)  , $\mathbf{F}_{\protect \overleftarrow{ji}}$ is the estimated optical flow from $I_i$ to $I_j$, and $C_x$ and $C_y$ are the estimated flow confidences in the $x$ and $y$ direction respectively. The first two rows correspond to static scenes where only the camera moves resulting in uniform flow across the image. The third row shows an example of a dynamic scene}
		\label{fig:tum_flow}
	\end{center}
    \vspace*{-3em}
\end{figure}

\begin{figure}[h!]
	\begin{center}
    	\includegraphics[width=\linewidth]{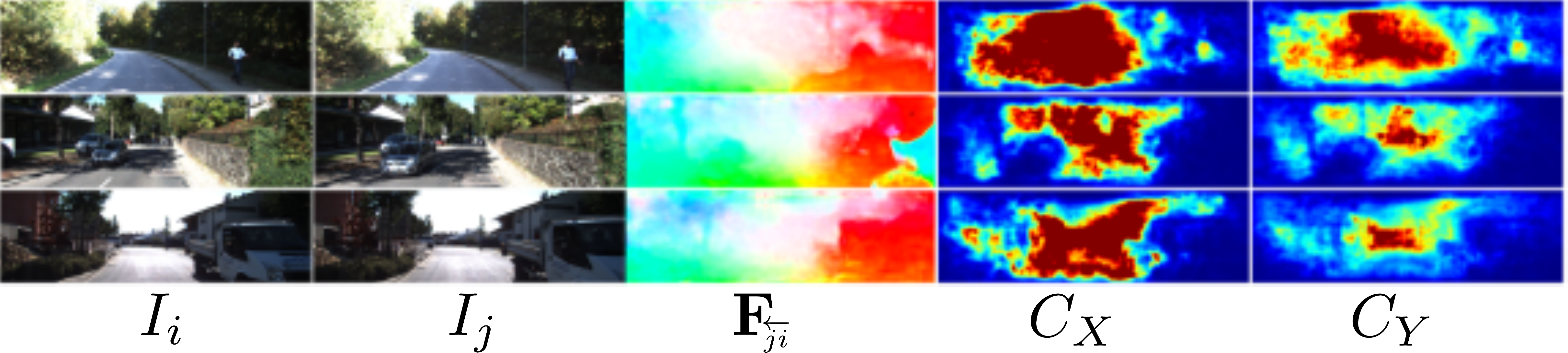}
		\caption{A selection of optical flow predictions made by our framework on the KITTI dataset\cite{Geiger2013IJRR}. Dynamic objects and the objects that do not appear in both frames due to large camera motion have low confidence}
		\label{fig:kitti_flow}
	\end{center}
    \vspace*{-3em}
\end{figure}

\begin{figure}[h!]
	\begin{center}
    	\includegraphics[width=0.20\linewidth]{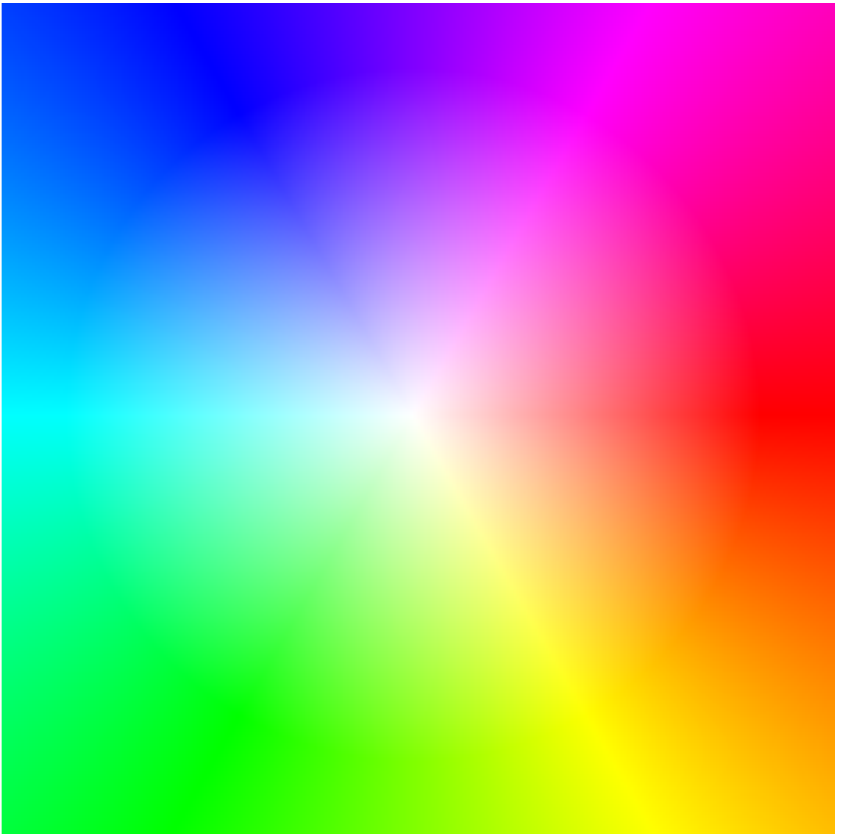}
		\caption{Optical flow color coding}
		\label{fig:flow_legend}
	\end{center}
    \vspace*{-3em}
\end{figure}
\clearpage
\pagebreak
\section{Network Architectures}
\subsection{Depth Network}
\begin{figure}[h!]
	\begin{center}
    	\includegraphics[width=\linewidth]{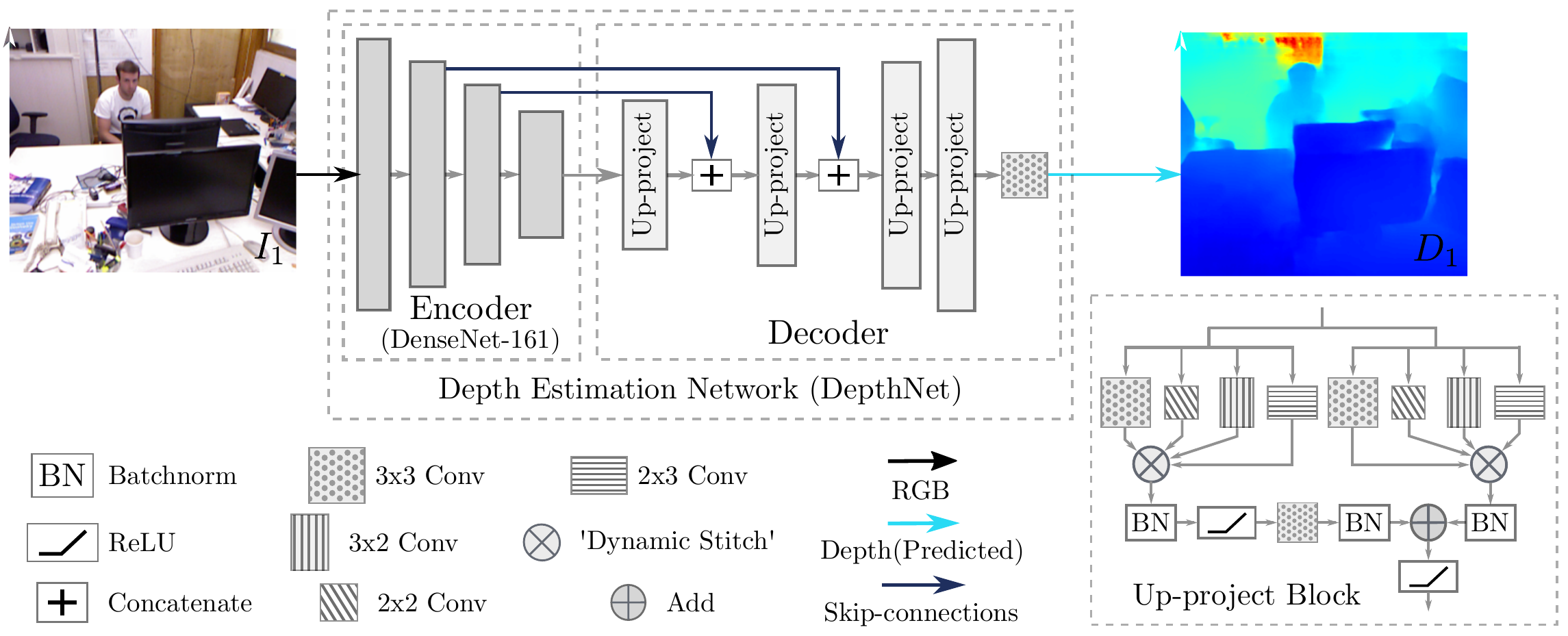}
		\caption{The Depth Prediction Network. We include a summary of all operations (\emph{bottom-left}) as well as description of the up-project blocks used in the decoder (\emph{bottom-right})}
		\label{fig:depth_net}
	\end{center}
 \end{figure}   
The encoder takes a global mean subtracted RGB image as an input, during the feature encoding stage the resolution of the activations are reduced by a factor of 16. First downsampling operation is performed using a strided convolutional layer, the next with a max-pooling layer and the final two with average-pooling layers. Up sampling process is performed using the up-project blocks proposed in \cite{laina2016}. Since the first down-sampling operation is performed by the very first convolutional layer and closely resemble image features,these activations are not provided to the decoder via a skip connection. It should be noted that ours isn't the first piece of work to predict depth using a DenseNet architecture. Kendall \emph{et al}. \cite{kendall2017uncertainties} also used a DenseNet variant and the gains that we obtain are predominantly due to the loss functions we employed. Appendix I shows the full breakdown of the architecture

\clearpage
\subsection{Flow Network}
\begin{figure}[h!]
	\begin{center}
    	\includegraphics[width=\linewidth]{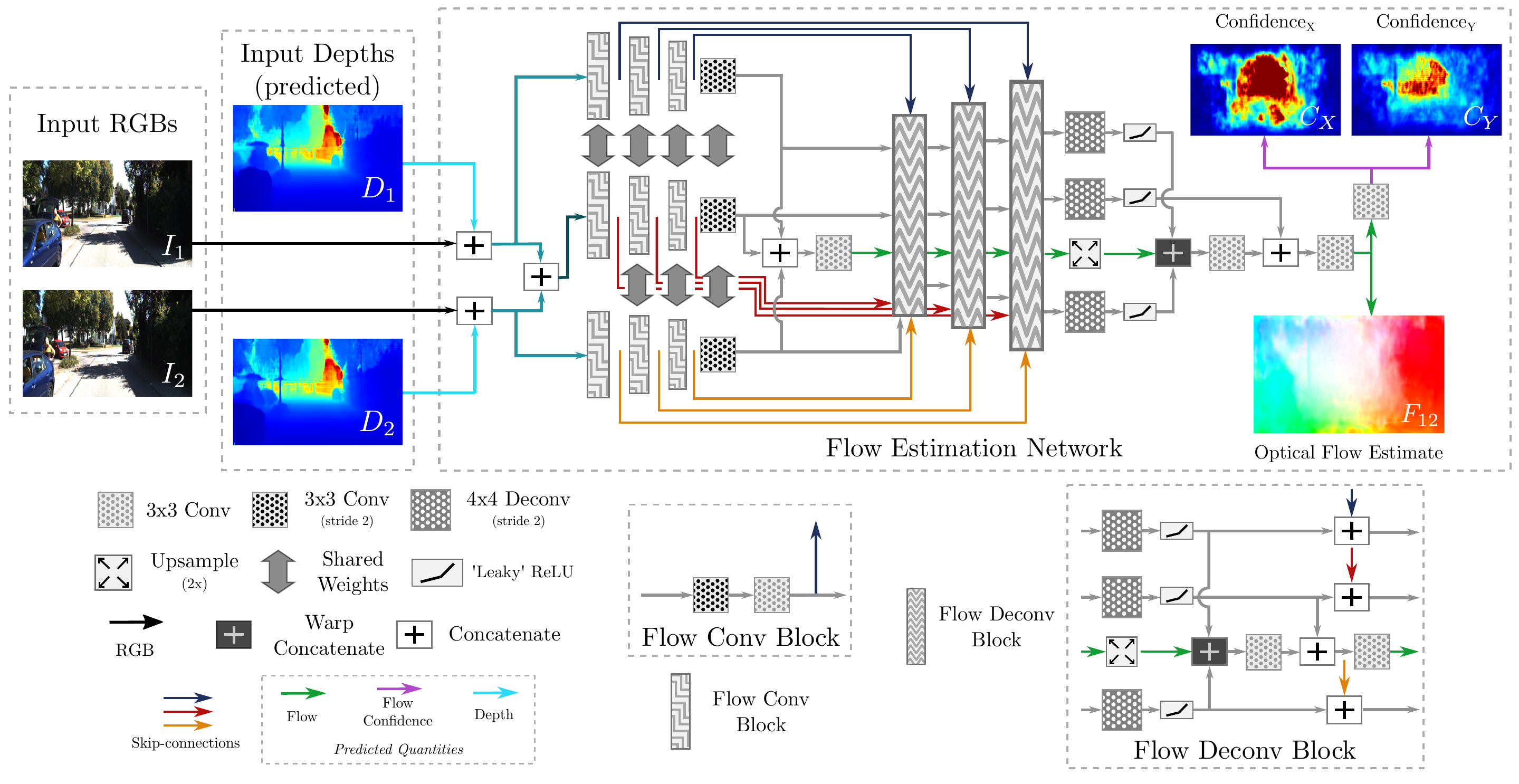}
		\caption{The optical flow prediction network. We include a summary of all operations (\emph{bottom-left}) as well as description of the \emph{flow-conv} and \emph{flow-deconv} blocks (\emph{bottom-right}) }
		\label{fig:nyu_worst}
	\end{center}
\end{figure}

The flow network has three streams. The first stream takes the left image and its' predicted depth map as an input, the second stream receives the right image and the corresponding predicted depth map and finally, the third stream receives both the left and right images and their associated depth predictions. Barring the first layer, all other layers of each stream share their weights. During the decoder stage the predicted flow is used to perform warp concatenations, where the right images activations are warped and concatenated with that of the left image. Since we are estimating optical flow in a coarse to fine manner, where the latter layers compute a residual to be added to the initial flow estimate, warp-concatenations help to capture the small displacements more effectively

\section{Pose Network}

\subsection{Iterative Re-weighted Approach}

As described in the main body of the paper, we are attempting to minimise the following error function with respect to the relative transformation parameters ($\xi_{\overleftarrow{21}} \in \mathbb{R}^6$):
\begin{equation}
\mathbf{e} = \sum_{i=1}^{N} ||(\mathbf{x}_i - \text{T}_{\overleftarrow{21}}\mathbf{x}_i)_{[u,v]} - \mathbf{F}_{\overleftarrow{21}}(\mathbf{u}_i)||_2 = \sum_{i=1}^{N} ||\mathbf{F}^+_{\overleftarrow{21}}(\mathbf{u}_i) - \mathbf{F}_{\overleftarrow{21}}(\mathbf{u}_i)||_2 = \sum_{i=1}^{N} \mathbf{r}_i^2,
\label{eq:error_function_iterative}
\end{equation}
For simplicity, we express the values in terms of normalised camera coordinates. The estimated flow $\mathbf{F}^+_{\overleftarrow{21}}$ is computed from the normalised camera coordinate and the current estimated transformation $\text{T}_{\overleftarrow{21}}\in \mathbb{SE}(3)$ as shown in Equation \ref{eq:error_function_iterative}. To simplify the mathematics we can represent the transformation using a matrix exponential as $\text{T}_{\overleftarrow{21}} = e^{\sum_{j=0}^{6}\alpha_j \mathbf{G}_j}$, where $\alpha_j \in \xi_{\overleftarrow{21}}$ is the $j^{th}$ component of the motion vector $\xi_{\overleftarrow{21}} \in \mathbb{R}^6$, which is a member of the Lie-algebra $\mathfrak{se}_3$, and $\mathbf{G}_j$ is the generator matrix corresponding to the relevant motion parameter. We can now differentiate the residual function with respect to the motion parameters to generate the following Jacobian
\begin{equation}
\text{J}_i=
\left[\begin{array}{cccccc}
\phantom{-}q & \phantom{-}0 & \phantom{1^2}-uq & \phantom{1^2}-uv & \phantom{1^2}u^2+1 & \phantom{1^2}-v \\
\phantom{-}0 & \phantom{-}q & \phantom{1^2}-vq & \phantom{1^2}-v^2-1 & \phantom{1^2}uv & \phantom{1^2}u \\
\end{array}\right],
\end{equation}
where $\text{J}_i$ is the $i^{th}$ Jacobian, which can be stacked to form a larger Jacobian matrix $\text{J} = \left[\text{J}_1^T, \text{J}_2^T, \dots, \text{J}_N^T\right]^T$, additionally the residual vectors can be stacked $\mathbf{r} = \left[\mathbf{r}^T_1, \mathbf{r}^T_2, \dots, \mathbf{r}^T_N\right]^T$. This allows us to iteratively reduce the loss function $\mathbf{e}$ using a standard Gauss-Newton approach given by
\begin{equation}
\beta = \left(\mathbf{J}^T\mathbf{W}\mathbf{J}\right)^{-1}\mathbf{J}^T\mathbf{W}\mathbf{r},
\end{equation}
where $\beta$ is the additive update to the motion parameters $\xi_{\overleftarrow{21}}$, and $\text{W}$ is a diagonal weight matrix $\text{W} = \text{diag}(\{\text{W}_1, \text{W}_2, \cdots, \text{W}_N\})$. $\text{W}_i = \text{W}(\mathbf{u_i})$ is the $i^{th}$ weight matrix defined by
\begin{equation}
\text{W}_i=
\left[\begin{array}{cc}
(C_{X}(\mathbf{u_i}) m^2) / (m^2 + r_{x}(\mathbf{u_i})^2) & 0 \\
0 & (C_{Y}(\mathbf{u_i}) m^2)/(m^2 + r_{y}(\mathbf{u_i})^2) \\
\end{array}\right],
\end{equation}
where $C_X(\mathbf{u_i})$ is the $i^{th}$ confidence value in the x-direction, $m$ is a constant that is computed from the residual $\mathbf{r}_i$ (Equation \ref{eq:error_function_iterative}), to be the mean residual magnitude of a single image, and $\mathbf{r}_X(\mathbf{u}_i)$ is the $i^{th}$ residual in the x-direction. This pipeline is implemented in Tensorflow \cite{abadi2016tensorflow} and allows us to train the network end to end.

\subsection{Network Based Approach}

This section was addressed in detail in the main body.
\pagebreak

\pagebreak
\clearpage
\section{Training Procedure}

\subsection{Depth Training}

All of the DenseNet-161 layers \cite{huang2017densely} of the depth nets are initialised using Imagenet\cite{ILSVRC15} pretrained weights. Remainder of the layers are intialised using MSRC\cite{he2015delving} initialisation. NYUv2\cite{Silberman} and TUM\cite{sturm12iros} models are trained purely using the supervised loss term. The network is regularized using a weight decay of 1$e^{-4}$ through out training and the learning rate schedule is shown below :

\begin{figure}[h]
	\begin{center}
    	\includegraphics[width=0.9\linewidth]{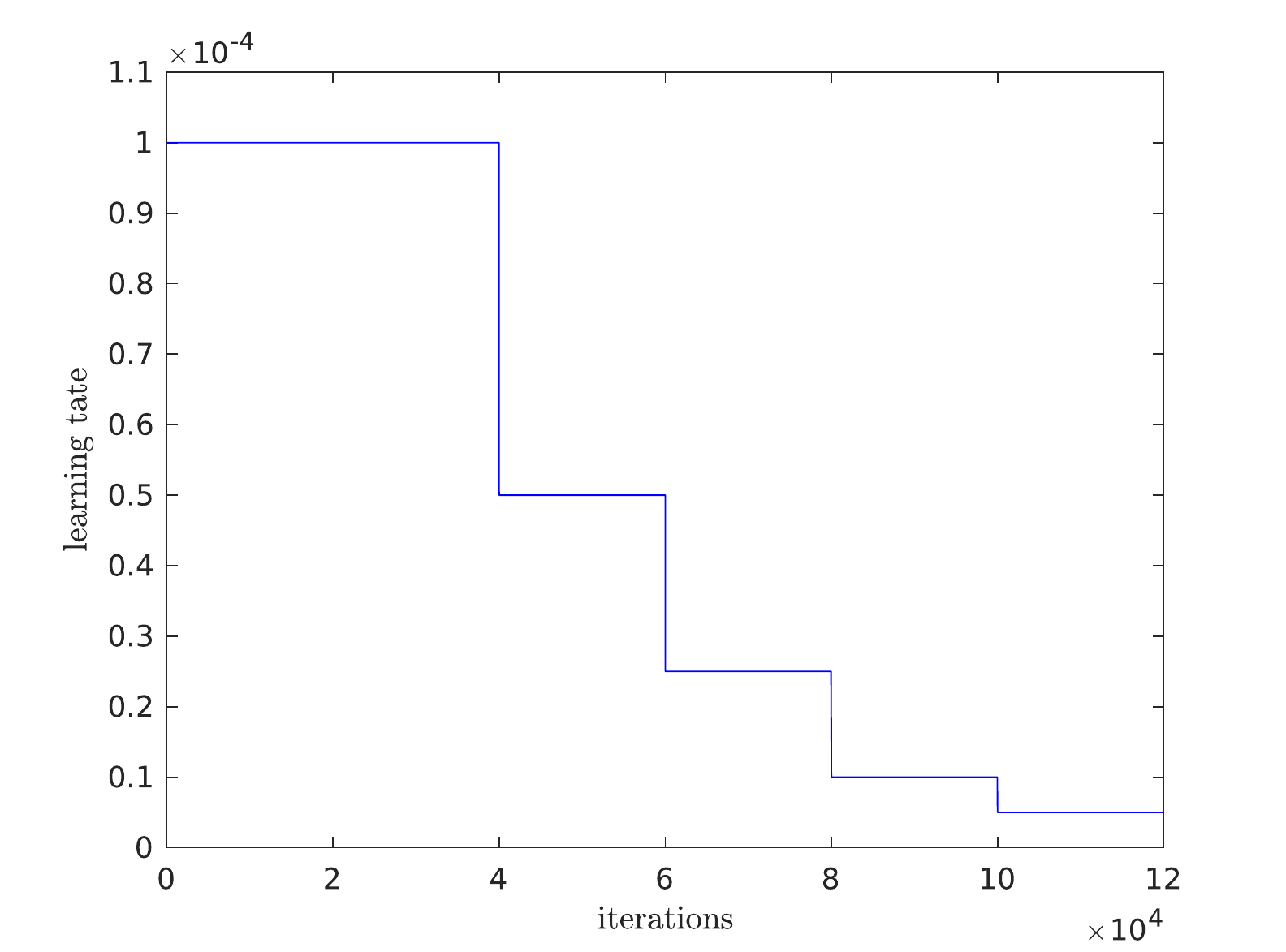}
		\caption{Learning rate schedule for NYUv2\cite{Silberman} depth training}
	\end{center}
\end{figure}

Out of $\approx$ 400,0000 images in the NYU dataset, we only use 12,000 during training. We perform data augmentation 4 times (a total training set of 48000 images) using color shifts, random crops and left-right flips. Although, data augmentation can be implemented during training we noticed a considerable speedup by performing data augmentation offline. The training images and the corresponding ground truth are downsampled by a factor of 2. Hence, the resolution of each training example becomes 320$\times$240. Each training batch contains 8 images and we use 4 GPUs, resulting in a overall batch size of 32. In terms of training speed we observe on average 19.3 examples/sec or 0.415 sec/batch. 

For the KITTI dataset we use 10,000 training images. Out of the training images that were defined in \cite{EigenNIPS} we further prune our training set to exclude any images that are part of the odometry test set. We adopt a learning rate schedule which spans for half the duration of the NYU. This is primarily to avoid over fitting as we are now working with a comparatively small training dataset.

\subsubsection{Optical Flow Training}

\begin{figure}[h]
	\begin{center}
    	\includegraphics[width=0.9\linewidth]{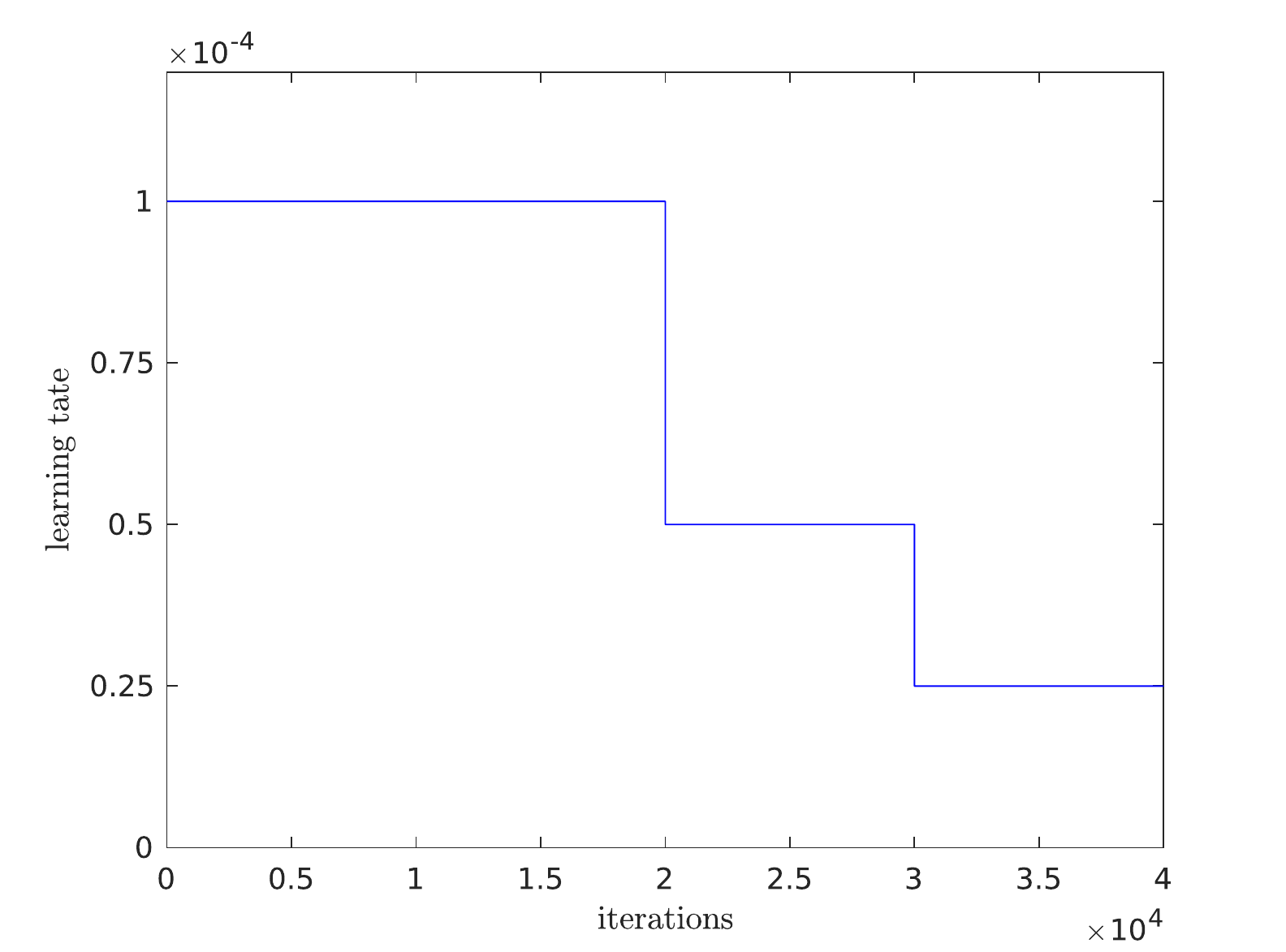}
		\caption{Learning rate schedule for optical flow training}
        \label{fig:flow_indoor}
	\end{center}
\end{figure}

In order to compute the ground truth optical flow image, for the NYUv2 \cite{Silberman} dataset we first compute the camera pose using the Iterative Closest Point (ICP) algorithm which can then be used with the ground truth depth map to compute optical flow. This process is slightly simplified for the TUM\cite{sturm12iros} dataset as the ground truth pose is provided. The network is then trained using the optical flow loss criterion. All the layers of the flow network are initialised using the MSRC\cite{he2015delving} initialisation and the learning rate schedule is shown in Figure \ref{fig:flow_indoor}. As it can be seen, the training duration is much smaller compared to the depth network training as the primary objective at this stage is to obtain a crude representation for both optical flow and the information matrix. Complete end-to-end fine tuning happens when the network is trained using the pose loss criterion.

\subsubsection{Pose Training}

We optimize the full network end-to-end using the pose loss and demonstrate that the state-of-the art depths can be further improved using the knowledge of pose. We train the network for 20,000 iterations with an initial learning rate of $1e^{-5}$ which is halved at the half-way point.

\clearpage
\section*{Appendix I}

We include a full table description of our depth network, for completeness.

\subsection*{Depth Net}

\begin{longtable}{ccccc}
        \multicolumn{4}{l}{Model Architechture Breakdown} \\ \toprule
         Layer & Channesl (I/O) & Scaling & Inputs \\ \midrule
         conv1  & 3/96 & 2 & Input Image \\
         pool1  & 96/96 & 4 & conv1 \\
         conv2\_1\_x1  & 96/192 & 4 & pool1 \\
         conv2\_1\_x2  & 192/48 & 4 &  conv2\_1\_x1 \\ 
         concat2\_1 & 144/144 & 4 &  conv2\_1\_x2, pool1 \\ 
         DenseBlk\_1  & 144/48 & 4 &  concat2\_1  \\ 
         concat2\_2  & 192/192 & 4 &  DenseBlk\_1, concat2\_1 \\ 
         DenseBlk\_2  & 192/48 & 4 &  concat2\_2  \\ 
         concat2\_3  & 240/240 & 4 &  DenseBlk\_2, concat2\_2 \\ 
         DenseBlk\_3  & 240/48 & 4 &  concat2\_3  \\ 
         concat2\_4  & 288/288 & 4 &  DenseBlk\_3, concat2\_3 \\ 
         DenseBlk\_4  & 288/48 & 4 &  concat2\_4  \\ 
         concat2\_5  & 336/336 & 4 &  DenseBlk\_4, concat2\_4 \\ 
         DenseBlk\_5  & 336/48 & 4 &  concat2\_5  \\ 
         concat2\_6  & 384/384 & 4 &  DenseBlk\_5, concat2\_5 \\ 
         conv2\_blk  & 384/192 & 4 & concat2\_6 \\ \midrule
         pool2  & 192/192 & 2 & conv2\_blk\\
         DenseBlk\_6  & 192/48 & 8 &  pool2  \\ 
         concat3\_1  & 240/240 & 8 &  DenseBlk\_6, pool2 \\ 
         DenseBlk\_7  & 240/48 & 8 &  concat3\_1  \\ 
         concat3\_2  & 288/288 & 8 &  DenseBlk\_7, concat3\_1 \\ 
         DenseBlk\_8  & 288/48 & 8 &  concat3\_2  \\ 
         concat3\_3  & 336/336 & 8 &  DenseBlk\_8, concat3\_2 \\ 
         DenseBlk\_9  & 336/48 & 8 &  concat3\_4  \\ 
         concat3\_4  & 384/384 & 8 &  DenseBlk\_9, concat3\_3 \\ 
         DenseBlk\_10 & 384/48 & 8 &  concat3\_4  \\
         concat3\_5  & 432/432 & 8 &  DenseBlk\_10, concat3\_4 \\
         DenseBlk\_11  & 432/48 & 8 &  concat3\_5  \\ 
         concat3\_6  & 480/480 & 8 &  DenseBlk\_11, concat3\_5 \\ 
         DenseBlk\_12  & 480/48 & 8 &  concat3\_6  \\ 
         concat3\_7  & 528/528 & 8 &  DenseBlk\_12, concat3\_6 \\ 
         DenseBlk\_13  & 528/48 & 8 &  concat3\_7  \\ 
         concat3\_8  & 576/576 & 8 &  DenseBlk\_13, concat3\_7 \\ 
         DenseBlk\_14 & 576/48 & 8 &  concat3\_8  \\
         concat3\_9  & 624/624 & 8 &  DenseBlk\_14, concat3\_8 \\
         DenseBlk\_15  & 624/48 & 8 &  concat3\_9  \\ 
         concat3\_10  & 672/672 & 8 &  DenseBlk\_15, concat3\_9 \\ 
         DenseBlk\_16  & 672/48 & 8 &  concat3\_10  \\ 
         concat3\_11  & 720/720 & 8 &  DenseBlk\_16, concat3\_10 \\ 
         DenseBlk\_17  & 720/48 & 8 &  concat3\_11  \\ 
         concat3\_12  & 768/768 & 8 &  DenseBlk\_17, concat3\_11 \\ 
         conv3\_blk  & 768/384 & 8 & concat3\_12 \\ \midrule
         pool3  & 384/384 & 16 & conv3\_blk\\
         DenseBlk\_18  & 384/48 & 8 &  pool3  \\ 
         concat4\_1  & 432/432 & 8 &  DenseBlk\_18, pool3 \\
         DenseBlk\_19 & 480/48  & 16 &  concat4\_2  \\
DenseBlk\_19 & 480/48  & 16 &  concat4\_2  \\
concat4\_2  & 528/528 & 16 &  DenseBlk\_19, concat4\_1 \\
DenseBlk\_20 & 528/48  & 16 &  concat4\_3  \\
concat4\_3  & 576/576 & 16 &  DenseBlk\_20, concat4\_2 \\
DenseBlk\_21 & 576/48  & 16 &  concat4\_4  \\
concat4\_4  & 624/624 & 16 &  DenseBlk\_21, concat4\_3 \\
DenseBlk\_22 & 624/48  & 16 &  concat4\_5  \\
concat4\_5  & 672/672 & 16 &  DenseBlk\_22, concat4\_4 \\
DenseBlk\_23 & 672/48  & 16 &  concat4\_6  \\
concat4\_6  & 720/720 & 16 &  DenseBlk\_23, concat4\_5 \\
DenseBlk\_24 & 720/48  & 16 &  concat4\_7  \\
concat4\_7  & 768/768 & 16 &  DenseBlk\_24, concat4\_6 \\
DenseBlk\_25 & 768/48  & 16 &  concat4\_8  \\
concat4\_8  & 816/816 & 16 &  DenseBlk\_25, concat4\_7 \\
DenseBlk\_26 & 816/48  & 16 &  concat4\_9  \\
concat4\_9  & 864/864 & 16 &  DenseBlk\_26, concat4\_8 \\
DenseBlk\_27 & 864/48  & 16 &  concat4\_10  \\
concat4\_10  & 912/912 & 16 &  DenseBlk\_27, concat4\_9 \\
DenseBlk\_28 & 912/48  & 16 &  concat4\_11  \\
concat4\_11  & 960/960 & 16 &  DenseBlk\_28, concat4\_10 \\
DenseBlk\_29 & 960/48  & 16 &  concat4\_12  \\
concat4\_12  & 1008/1008 & 16 &  DenseBlk\_29, concat4\_11 \\
DenseBlk\_30 & 1008/48  & 16 &  concat4\_13  \\
concat4\_13  & 1056/1056 & 16 &  DenseBlk\_30, concat4\_12 \\
DenseBlk\_31 & 1056/48  & 16 &  concat4\_14  \\
concat4\_14  & 1104/1104 & 16 &  DenseBlk\_31, concat4\_13 \\
DenseBlk\_32 & 1104/48  & 16 &  concat4\_15  \\
concat4\_15  & 1152/1152 & 16 &  DenseBlk\_32, concat4\_14 \\
DenseBlk\_33 & 1152/48  & 16 &  concat4\_16  \\
concat4\_16  & 1200/1200 & 16 &  DenseBlk\_33, concat4\_15 \\
DenseBlk\_34 & 1200/48  & 16 &  concat4\_17  \\
concat4\_17  & 1248/1248 & 16 &  DenseBlk\_34, concat4\_16 \\
DenseBlk\_35 & 1248/48  & 16 &  concat4\_18  \\
concat4\_18  & 1296/1296 & 16 &  DenseBlk\_35, concat4\_17 \\
DenseBlk\_36 & 1296/48  & 16 &  concat4\_19  \\
concat4\_19  & 1344/1344 & 16 &  DenseBlk\_36, concat4\_18 \\
DenseBlk\_37 & 1344/48  & 16 &  concat4\_20  \\
concat4\_20  & 1392/1392 & 16 &  DenseBlk\_37, concat4\_19 \\
DenseBlk\_38 & 1392/48  & 16 &  concat4\_21  \\
concat4\_21  & 1440/1440 & 16 &  DenseBlk\_38, concat4\_20 \\
DenseBlk\_39 & 1440/48  & 16 &  concat4\_22  \\
concat4\_22  & 1488/1488 & 16 &  DenseBlk\_39, concat4\_21 \\
DenseBlk\_40 & 1488/48  & 16 &  concat4\_23  \\
concat4\_23  & 1536/1536 & 16 &  DenseBlk\_40, concat4\_22 \\
DenseBlk\_41 & 1536/48  & 16 &  concat4\_24  \\
concat4\_24  & 1584/1584 & 16 &  DenseBlk\_41, concat4\_23 \\
DenseBlk\_42 & 1584/48  & 16 &  concat4\_25  \\
concat4\_25  & 1632/1632 & 16 &  DenseBlk\_42, concat4\_24 \\
DenseBlk\_43 & 1632/48  & 16 &  concat4\_26  \\
concat4\_26  & 1680/1680 & 16 &  DenseBlk\_43, concat4\_25 \\
DenseBlk\_44 & 1680/48  & 16 &  concat4\_27  \\
concat4\_27  & 1728/1728 & 16 &  DenseBlk\_44, concat4\_26 \\
DenseBlk\_45 & 1728/48  & 16 &  concat4\_28  \\
concat4\_28  & 1776/1776 & 16 &  DenseBlk\_45, concat4\_27 \\
DenseBlk\_46 & 1776/48  & 16 &  concat4\_29  \\
concat4\_29  & 1824/1824 & 16 &  DenseBlk\_46, concat4\_28 \\
DenseBlk\_47 & 1824/48  & 16 &  concat4\_30  \\
concat4\_30  & 1872/1872 & 16 &  DenseBlk\_47, concat4\_29 \\
DenseBlk\_48 & 1872/48  & 16 &  concat4\_31  \\
concat4\_31  & 1920/1920 & 16 &  DenseBlk\_48, concat4\_30 \\
DenseBlk\_49 & 1920/48  & 16 &  concat4\_32  \\
concat4\_32  & 1968/1968 & 16 &  DenseBlk\_49, concat4\_31 \\
DenseBlk\_50 & 1968/48  & 16 &  concat4\_33  \\
concat4\_33  & 2016/2016 & 16 &  DenseBlk\_50, concat4\_32 \\
DenseBlk\_51 & 2016/48  & 16 &  concat4\_34  \\
concat4\_34  & 2064/2064 & 16 &  DenseBlk\_51, concat4\_33 \\
DenseBlk\_52 & 2064/48  & 16 &  concat4\_35  \\
concat4\_35  & 2112/2112 & 16 &  DenseBlk\_52, concat4\_34 \\
DenseBlk\_53 & 2112/48  & 16 &  concat4\_36  \\
concat4\_36  & 2160/2160 & 16 &  DenseBlk\_53, concat4\_35 \\
conv4\_blk  & 2160/1056 & 16 & concat4\_36 \\ \midrule
DenseBlk\_54  & 1056/48 & 16 &  conv4\_blk  \\ 
concat5\_1  & 1104/1104 & 16 &  conv4\_blk, DenseBlk\_54 \\
DenseBlk\_55 & 1104/48  & 16 &  concat5\_2  \\
concat5\_2  & 1152/1152 & 16 &  DenseBlk\_55, concat5\_1 \\
DenseBlk\_56 & 1152/48  & 16 &  concat5\_3  \\
concat5\_3  & 1200/1200 & 16 &  DenseBlk\_56, concat5\_2 \\
DenseBlk\_57 & 1200/48  & 16 &  concat5\_4  \\
concat5\_4  & 1248/1248 & 16 &  DenseBlk\_57, concat5\_3 \\
DenseBlk\_58 & 1248/48  & 16 &  concat5\_5  \\
concat5\_5  & 1296/1296 & 16 &  DenseBlk\_58, concat5\_4 \\
DenseBlk\_59 & 1296/48  & 16 &  concat5\_6  \\
concat5\_6  & 1344/1344 & 16 &  DenseBlk\_59, concat5\_5 \\
DenseBlk\_60 & 1344/48  & 16 &  concat5\_7  \\
concat5\_7  & 1392/1392 & 16 &  DenseBlk\_60, concat5\_6 \\
DenseBlk\_61 & 1392/48  & 16 &  concat5\_8  \\
concat5\_8  & 1440/1440 & 16 &  DenseBlk\_61, concat5\_7 \\
DenseBlk\_62 & 1440/48  & 16 &  concat5\_9  \\
concat5\_9  & 1488/1488 & 16 &  DenseBlk\_62, concat5\_8 \\
DenseBlk\_63 & 1488/48  & 16 &  concat5\_10  \\
concat5\_10  & 1536/1536 & 16 &  DenseBlk\_63, concat5\_9 \\
DenseBlk\_64 & 1536/48  & 16 &  concat5\_11  \\
concat5\_11  & 1584/1584 & 16 &  DenseBlk\_64, concat5\_10 \\
DenseBlk\_65 & 1584/48  & 16 &  concat5\_12  \\
concat5\_12  & 1632/1632 & 16 &  DenseBlk\_65, concat5\_11 \\
DenseBlk\_66 & 1632/48  & 16 &  concat5\_13  \\
concat5\_13  & 1680/1680 & 16 &  DenseBlk\_66, concat5\_12 \\
DenseBlk\_67 & 1680/48  & 16 &  concat5\_14  \\
concat5\_14  & 1728/1728 & 16 &  DenseBlk\_67, concat5\_13 \\
DenseBlk\_68 & 1728/48  & 16 &  concat5\_15  \\
concat5\_15  & 1776/1776 & 16 &  DenseBlk\_68, concat5\_14 \\
DenseBlk\_69 & 1776/48  & 16 &  concat5\_16  \\
concat5\_16  & 1824/1824 & 16 &  DenseBlk\_69, concat5\_15 \\
DenseBlk\_70 & 1824/48  & 16 &  concat5\_17  \\
concat5\_17  & 1872/1872 & 16 &  DenseBlk\_70, concat5\_16 \\
DenseBlk\_71 & 1872/48  & 16 &  concat5\_18  \\
concat5\_18  & 1920/1920 & 16 &  DenseBlk\_71, concat5\_17 \\
DenseBlk\_72 & 1920/48  & 16 &  concat5\_19  \\
concat5\_19  & 1968/1968 & 16 &  DenseBlk\_72, concat5\_18 \\
DenseBlk\_73 & 1968/48  & 16 &  concat5\_20  \\
concat5\_20  & 2016/2016 & 16 &  DenseBlk\_73, concat5\_19 \\
DenseBlk\_74 & 2016/48  & 16 &  concat5\_21  \\
concat5\_21  & 2064/2064 & 16 &  DenseBlk\_74, concat5\_20 \\
DenseBlk\_75 & 2064/48  & 16 &  concat5\_22  \\
concat5\_22  & 2112/2112 & 16 &  DenseBlk\_75, concat5\_21 \\
DenseBlk\_76 & 2112/48  & 16 &  concat5\_23  \\
concat5\_23  & 2160/2160 & 16 &  DenseBlk\_76, concat5\_22 \\
DenseBlk\_77 & 2160/48  & 16 &  concat5\_24  \\
concat5\_24  & 2208/2208 & 16 &  DenseBlk\_77, concat5\_23 \\
conv5\_blk  & 2208/1024 & 16 & concat5\_24  \\ \midrule
upproject\_1  & 1024/512 & 8 & concat5\_24  \\ \midrule
concat\_up\_2 & 896/896 & 8 & upproject\_1, conv3\_blk \\
upproject\_2  & 896/584 & 4 & concat\_up\_2  \\ \midrule
concat\_up\_3 & 776/776 & 4 & upproject\_2, conv2\_blk \\
upproject\_3  & 776/256 & 2 & concat\_up\_3  \\ \midrule
upproject\_4  & 256/128 & 1 & upproject\_3 \\ \midrule
conv\_pred	& 128/1 & 1 & upproject\_4 


\label{tab:model_arch}    
\end{longtable}

\bibliographystyle{splncs04}

\end{document}